\documentclass[conference]{IEEEtran}
\IEEEoverridecommandlockouts

\usepackage{cite}
\usepackage{amsmath,amssymb,amsfonts}
\usepackage{algorithmic}
\usepackage{graphicx}
\usepackage{textcomp}
\usepackage{xcolor}

\usepackage[hyphens]{url}
\usepackage{booktabs}
\usepackage{multirow}
\usepackage{array}
\usepackage{colortbl}
\usepackage{tikz}
\usepackage{algorithm}
\usepackage{placeins}
\usepackage{stfloats}
\makeatletter
\g@addto@macro\normalsize{%
\setlength{\abovedisplayskip}{2pt plus 1pt minus 1pt}%
\setlength{\belowdisplayskip}{2pt plus 1pt minus 1pt}%
\setlength{\abovedisplayshortskip}{1pt plus 1pt}%
\setlength{\belowdisplayshortskip}{1pt plus 1pt}}
\makeatother

\usepackage{CJKutf8}
\usepackage[normalem]{ulem}
\usepackage[hidelinks]{hyperref}

\let\citep\cite

\title{Beyond Fixed Prompt Configuration: Alternating Policy Learning for Adaptive Topology-Based Prompting in Few-Shot Graph Learning}
\author{\IEEEauthorblockN{Yizhuo Ma\textsuperscript{1}, Yihong Huang\textsuperscript{1}, Rongzheng Wang\textsuperscript{1}, Jiakai Li\textsuperscript{1}, Ke Qin\textsuperscript{1}, Shuang Liang\textsuperscript{1*}, Jie Shao\textsuperscript{1}}
\IEEEauthorblockA{\textsuperscript{1}University of Electronic Science and Technology of China, Chengdu, China\\
myz@std.uestc.edu.cn, hyh@std.uestc.edu.cn, wangrongzheng@std.uestc.edu.cn, ljk@std.uestc.edu.cn,\\
 qinke@uestc.edu.cn, shuangliang@uestc.edu.cn, shaojie@uestc.edu.cn}\thanks{\textsuperscript{*}Shuang Liang is the corresponding author.\newline\hspace*{\parindent}\protect\hypertarget{code-availability}{\textsuperscript{\ensuremath{\dagger}}}The code is available at \url{https://github.com/myz12138/TTPR}.}}
\begin{document}
\begin{CJK*}{UTF8}{gbsn}

\maketitle
\setlength{\abovedisplayskip}{2pt plus 1pt minus 1pt}
\setlength{\belowdisplayskip}{2pt plus 1pt minus 1pt}
\setlength{\abovedisplayshortskip}{1pt plus 1pt}
\setlength{\belowdisplayshortskip}{1pt plus 1pt}

\begin{abstract}

Graph Prompt Learning (GPL) adapts frozen pre-trained graph models to few-shot graph tasks through a small set of trainable prompts that encode task-specific information. Topology-based prompting instantiates this idea by augmenting an input graph with new prompt nodes and edges, yielding promising performance. However, prompt content and propagation structure are coupled, making it difficult to assess their individual effects and learn compatible configurations. In this paper, we empirically analyze two challenges in learning adaptive policies for topology-based prompting: asymmetric performance effects and coupled configuration selection. We then propose TTPR, a two-stage reinforcement framework that alternately optimizes two policies to generate adaptive configurations, with each policy conditioned on configurations generated by the other fixed policy.
These conditional updates separate policy evaluation while preserving configuration compatibility. Specifically, TTPR models an edge-level policy to construct connection patterns and optimizes it with Connection Relative Policy Optimization (CRPO) through group-wise comparisons of complete connection patterns. The other is a node-level policy that generates prompt-node schemes and is optimized with Incremental Node Proximal Policy Optimization (INPPO) through incremental prompt-count decisions and feature generation. Across diverse datasets and pre-training strategies, TTPR ranks first in 8/10 few-shot node classification settings and attains the highest average performance across graph classification benchmarks\hyperlink{code-availability}{\textsuperscript{\ensuremath{\dagger}}}.

\end{abstract}

\begin{figure*}[!t]
    \centering
    \includegraphics[width=1.0\textwidth]{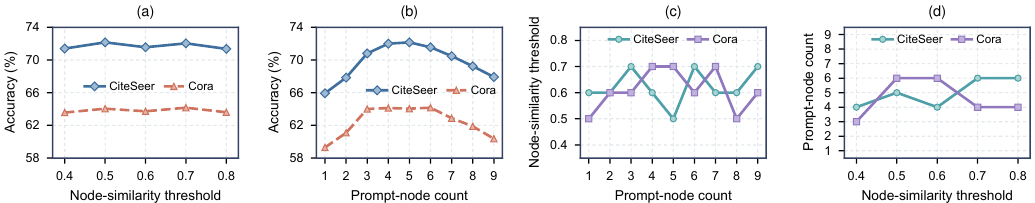}
    \par
\caption{Observational experiments on topology-based prompting configurations for 10-shot node classification on Cora and CiteSeer with a frozen, GPPT-pretrained GNN. The node-similarity threshold specifies the minimum feature similarity required to connect a prompt node with an input node, while the prompt-node count denotes the number of auxiliary prompt nodes. For each configuration, we select the checkpoint with the highest validation accuracy and report its test accuracy. Figures (a--b) fix the prompt-node count at 5 and the node-similarity threshold at 0.5, respectively, and vary the other parameter. Figures (c--d) fix the parameter on the horizontal axis and select the best value of the other parameter based on validation accuracy.}
    \label{fig:intro-observations}
\end{figure*}

\section{Introduction}

Graph pre-training captures transferable knowledge from unlabeled data, yet adapting this knowledge to downstream tasks remains challenging when annotations are scarce~\citep{pre-train1,lowdata1}. Graph Prompt Learning (GPL) addresses this challenge by adapting frozen pretrained models through a small set of learnable prompts~\citep{GPL1,GPL2,lowdata3}. These prompts modify graph features or structures and are optimized using downstream labels, guiding the frozen encoder to produce task-relevant representations. Most approaches adopt the Prompt as Tokens (PT) paradigm, injecting prompt vectors into node attributes or subgraph representations~\citep{Prompt-as-tokens}. GPF~\citep{GPF} learns a shared feature prompt, while RELIEF~\citep{RELIEF} uses reinforcement learning to tailor prompt placement and content to each input. Such feature-level adaptation controls the information supplied by prompts, but its propagation remains constrained by the original graph connectivity.

Prompt as Graphs (PG) expands this design space by introducing auxiliary nodes and connections that provide additional paths for task-relevant information~\citep{Prompt-as-graph}. This topology-based prompting combines a prompt-node scheme that specifies prompt content with a connection pattern that governs its propagation. For example, All-in-One~\citep{ALL-in-One} constructs a graph of learnable prompt nodes and attaches it to the input graph using feature-similarity thresholds. PSP~\citep{PSP2024} further introduces class prototypes as virtual nodes and learns their connection weights to original nodes, improving prototype estimation under limited supervision. These methods focus on learning prompt features or connections within predefined configuration schemes, leaving the coordinated adaptation of prompt-node counts, features, and connection patterns underexplored. Our empirical analysis reveals two challenges in coordinating these configurations.

\textbf{Asymmetric performance effects.} Different prompt-node schemes introduce distinct prompt information into the graph, whereas different connection patterns produce distinct propagation patterns, causing downstream performance to vary at different scales. Figure~\ref{fig:intro-observations}(a--b) shows that connection patterns constructed using different node-similarity thresholds produce substantially smaller performance variations than prompt-node schemes with different prompt-node counts across both datasets.  Simultaneous changes to both configurations confound their contributions to the shared performance feedback, allowing larger node-level effects to overshadow gains from connection refinement.

\textbf{Coupled configuration selection.} Effective topology-based prompting requires compatible prompt-node schemes and connection patterns, as different pairings yield different downstream outcomes. Figure~\ref{fig:intro-observations}(c--d) substantiates this requirement, showing that a configuration value that is optimal in one pairing may not remain optimal in another, resulting in different best-performing combinations.  Prompt counts and features shape the choice of connections, whose propagation effects in turn determine the utility of the prompt-node scheme. Independent selection can therefore produce incompatible configurations. 

These observations show that effective policy learning requires distinguishable evaluation of the two policies while preserving configuration compatibility. To address this, we propose \textbf{TTPR}, a \textbf{T}wo-stage \textbf{T}opology-based \textbf{P}rompting \textbf{R}einforcement framework that separately models node-level and edge-level policies to learn prompt-node schemes and connection patterns, respectively. By keeping one policy fixed while optimizing the other, TTPR provides distinguishable performance feedback without breaking the compatibility between their generated configurations. In Stage~I, we model connection selection as a binary decision process over candidate prompt-related edges, with each action specifying a complete connection pattern. We propose \textbf{Connection Relative Policy Optimization (CRPO)}, which evaluates alternative patterns under a fixed prompt-node scheme and learns from their relative downstream rewards. Building on GRPO's group-relative advantage estimation~\citep{GRPO}, CRPO forms comparison groups by sampling graph connections at the input of the frozen GNN, allowing the edge policy to assess the collective propagation effects of its decisions under shared prompt content. In Stage~II, we model prompt-node configuration as an incremental decision process that adjusts the prompt count and generates corresponding features under a fixed edge policy. We propose \textbf{Incremental Node Proximal Policy Optimization (INPPO)}, which couples discrete count adjustments with conditional feature generation and learns these hybrid decisions from downstream feedback along configuration trajectories. INPPO adopts PPO's clipped policy objective~\citep{PPO} to jointly update the count and feature policies, allowing changes in prompt capacity to be evaluated together with the content they introduce. The two stages alternate until the iteration limit is reached. Our contributions are summarized as follows: 

\textbf{(1)} We conduct observational experiments on topology-based prompting configurations for few-shot graph learning and identify two key challenges: asymmetric performance effects and coupled configuration selection.

\textbf{(2)} We propose TTPR for adaptive topology-based prompting, which separately models node-level and edge-level policies and alternately optimizes them with CRPO and INPPO to learn compatible prompt configurations.

\textbf{(3)} Across diverse datasets and pre-training strategies, TTPR ranks first in 8/10 few-shot node classification settings and attains the highest average performance across graph classification benchmarks.

\section{Related Work}

\subsection{Graph Prompt Learning}

Graph prompt learning adapts pretrained graph models to downstream tasks through learnable prompts~\citep{PROMPT}. Prompt as Tokens represents prompts as learnable vectors applied to node attributes or representations. GPF~\citep{GPF} adds a shared prompt vector to node features, while SUPT~\citep{SUPT} assigns prompts at the subgraph level. SGL-PT~\citep{GPL4} bridges pre-training and graph classification through a task-compatible prompting function, and DAGPrompT~\citep{DAGPrompT} combines hop-specific prompts with low-rank adaptation to accommodate distribution shifts. Further work explores the role of edges in graph prompting. EdgePrompt~\citep{EdgePrompt} incorporates learnable edge prompt vectors into message passing~\citep{DSAS2026}. Subsequent studies adjust local connections through probabilistic~\citep{BetaLR2024} edge selection and differentiable relaxation~\citep{GraphTOP2025}, or learn edge weights on feature-similarity-based candidate connections and progressively integrate them with the original graph~\citep{UniPrompt2025}. Prompt as Graphs constructs prompt-augmented graphs by introducing prompt nodes and their connections. All-in-One~\citep{ALL-in-One} uses similarity-based prompt graphs, while PSP~\citep{PSP2024} learns connections to class-prototype nodes. Further studies combine cluster prompt nodes with meta-path information~\citep{NeuroPath2025} for heterogeneous graph adaptation~\citep{CLEAR2025}, or organize query attributes and virtual nodes into prompt graphs for attributed community search~\citep{PLACE2025}. These methods typically construct prompts according to predefined rules or task semantics~\citep{GraphToolInstruction2025} and optimize prompt parameters or connection weights through gradient-based learning. Recent work further investigates reinforcement learning for input-adaptive prompt construction. RELIEF~\citep{RELIEF} formulates prompt placement and content generation as sequential decisions, while LEAP~\citep{LEAP} adapts node prompts through selection and editing. Despite their effectiveness, these RL methods primarily configure prompts on existing nodes, with information propagation governed by the original connections~\citep{KEPo2026}. TTPR introduces auxiliary prompt structures, jointly configuring prompt-node counts, features, and connections through alternating node-level and edge-level policy learning to coordinate prompt content and propagation structure.

\subsection{Reinforcement Learning}
RL supports structured decision-making across combinatorial optimization, graph learning, and data-driven planning~\citep{TopoLLM2026,RL-survey-combination,GRL-survey,RL-survey}. Hybrid Actor--Critic methods address parameterized action spaces that combine discrete choices with continuous parameters~\citep{H-PPO}. Proximal Policy Optimization (PPO)~\citep{PPO} constrains policy updates with a clipped surrogate objective, while Group Relative Policy Optimization (GRPO)~\citep{GRPO} derives relative advantages from rewards within groups of sampled responses. RL has also supported foundation-model alignment through RL from human feedback~\citep{RLHF}, graph subset selection and pooling~\citep{GraphCogent2026,RL-for-graph1,RL-for-graph2}, and sequential node acquisition for active learning~\citep{RL-for-graph3}. These studies establish RL as an effective tool for structured graph decisions. Yet topology-based prompting combines combinatorial search over connection patterns with sequential node refinement, requiring policies tailored to these distinct action spaces.

\section{Preliminaries}

\subsection{Topology-based Prompting on Graph}
Let $\mathcal{G}=(\mathcal{V},\mathbf{X},\mathbf{A})$ denote the downstream graph, where $\mathcal{V}$ contains $N$ nodes, $\mathbf{X}\in\mathbb{R}^{N\times d}$ is the node feature matrix, and $\mathbf{A}\in\{0,1\}^{N\times N}$ is the adjacency matrix. A self-supervised pre-trained GNN $f_\theta$ remains frozen during adaptation. Topology-based prompting introduces $M$ learnable prompt nodes $\mathcal{U}=\{u_1,\dots,u_M\}$ with features $\mathbf{P}\in\mathbb{R}^{M\times d}$, together with cross-connections between prompt and original nodes and connections among prompt nodes. The resulting prompt-augmented graph $\mathcal{G}'=(\mathcal{V}',\mathbf{X}',\mathbf{A}')$ is defined as:
\begin{equation}
    \mathcal{V}'=\mathcal{V} \cup \mathcal{U},
    \
    \mathbf{X}' = \begin{bmatrix}
        \mathbf{X} \\
        \mathbf{P}
    \end{bmatrix} \in \mathbb{R}^{(N+M) \times d}, 
    \
    \mathbf{A}' = \begin{bmatrix}
        \mathbf{A} & {E}_{\text{op}} \\
        {E}_{\text{po}} & {E}_{\text{pp}}
    \end{bmatrix},
\end{equation}
For molecular graphs, $\mathbf X$ denotes the continuous atom embeddings produced by the frozen pretrained input embedding layer.
where ${E}_{\text{op}}\in\{0,1\}^{N\times M}$ and ${E}_{\text{po}}\in\{0,1\}^{M\times N}$ encode original--prompt cross-connections, and ${E}_{\text{pp}}\in\{0,1\}^{M\times M}$ encodes prompt--prompt connections.

With $f_\theta$ frozen, downstream adaptation jointly optimizes the prompt parameters $\phi$, comprising the prompt features $\mathbf P$ and the added edges, and the task classifier parameters $\psi$ by maximizing the joint likelihood of the correct labels over labeled samples:
\begin{equation}
\max_{\phi,\psi}\prod_{i\in\mathcal D_L}p\bigl(y_i\mid g_\psi\bigl(\mathcal R(f_\theta(\mathbf X',\mathbf A';\phi))\bigr)_i\bigr),
\end{equation}
where $\mathcal D_L$ denotes the set of labeled samples, $f_\theta$ is the frozen pre-trained graph encoder, $\mathcal R$ is a readout operator that averages the node representations within each input prompt-augmented graph, and $g_\psi$ is the task classifier. For node classification, the input is the prompt-augmented induced subgraph associated with the target node. For graph classification, the input is the entire prompt-augmented graph. The subscript $i$ identifies the corresponding sample, and $p(y_i\mid\cdot)$ is the predicted probability of its label $y_i$. Node classification uses cross-entropy loss, while molecular graph classification uses binary cross-entropy loss computed only over available labels. Both policy-learning stages use the negative task loss as the reward, and the task head is trained with the same task loss under supervision.

{

\subsection{Reinforcement Learning (RL) Basics}

RL learns sequential decisions through interactions with an environment. Formally, the environment is modeled as a Markov Decision Process (MDP) $(\mathcal{S}, \mathcal{A}, \pi_{\theta}, r)$, where $\mathcal{S}$, $\mathcal{A}$, $\pi_\theta$, and $r$ denote the state space, action space, state-conditioned policy parameterized by $\theta$, and reward function, respectively. At each discrete step $t=0,1,\dots$, the agent observes $s_t\in\mathcal{S}$, samples $a_t\in\mathcal{A}$ from $\pi_\theta$, receives a scalar reward $r_t$, and transitions to $s_{t+1}$, producing the trajectory:
\begin{equation}
\tau = (s_0, a_0, r_0, s_1, a_1, r_1, \dots, s_T),
\end{equation}
where $T$ is the trajectory length. In this paper, the state $s$ is a vector representation of the current graph and prompt configuration.
}

\section{Methodology}
\begin{figure*}[tbp]
    \centering
    \includegraphics[width=1.0\textwidth]{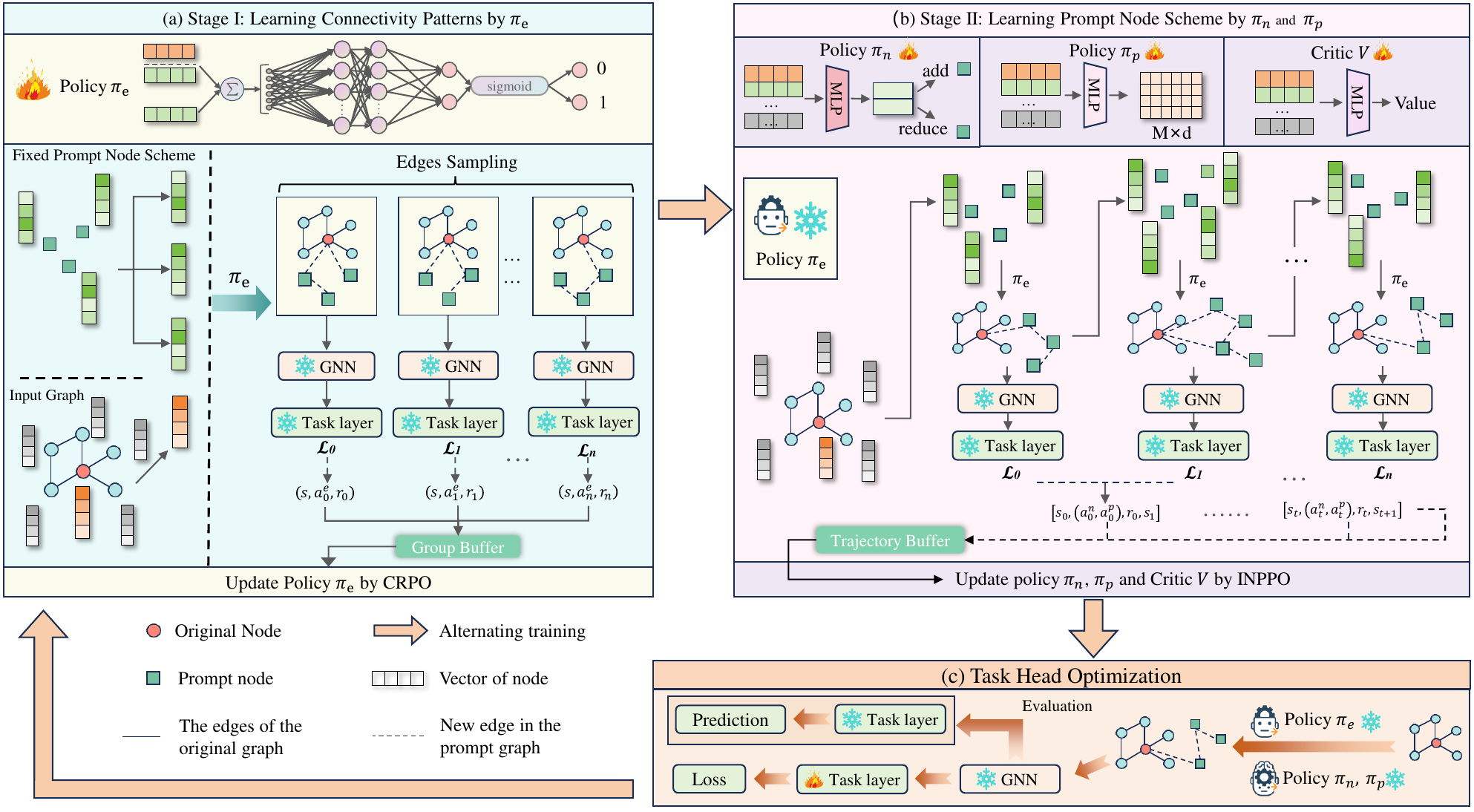}
    \caption{Overview of TTPR. Each cycle includes (a) edge-level policy learning, (b) node-level policy learning, and (c) task-head optimization. The loop stops at the iteration limit.}
    \label{fig:framework}
\end{figure*}

Figure~\ref{fig:framework} illustrates the alternating training process of TTPR. The framework comprises an edge-policy module containing $\pi_e$ and a node-policy module containing the prompt-count policy $\pi_n$ and feature-generation component $\pi_p$.  Stage~I learns connection patterns with the node-policy parameters fixed. Stage~II incrementally updates the prompt configuration with the edge-policy parameters fixed. The policy modules are then held fixed while the task head is optimized, followed by the next alternating cycle.

\subsection{Stage I: Edge-level Policy Learning}
\label{stage1}

\noindent\textbf{Motivation 1.} With a fixed prompt-node scheme, there are up to \mbox{$N_cM+\binom{M}{2}$} candidate undirected connections, yielding $2^{N_cM+\binom{M}{2}}$ possible connection patterns, where $N_c=1$ for node classification and $N_c=N$ for graph classification. Here, $N_c$ counts the original nodes eligible for prompt connections, namely the target node for node classification and all original nodes for graph classification. The number of possible patterns grows exponentially with the number of candidate connections. Moreover, Figure~\ref{fig:intro-observations}(a) shows that different connection patterns under a fixed prompt-node scheme often yield only small performance gaps, making it harder to compare their effects using absolute downstream rewards alone. We therefore represent each connection pattern through edge-level decisions and evaluate complete patterns through group-wise comparison. Specifically, multiple patterns are sampled from the same graph state by varying combinations of edges. Their prompt-augmented graphs are evaluated by the frozen GNN, and relative advantages derived from downstream performance provide discriminative feedback for optimizing these decisions.

\noindent\textbf{Edge-level Policy.} We define the edge-level policy $\pi_e$ over the action space $\mathcal A^e$, where each action specifies binary decisions for undirected prompt--original and prompt--prompt connections. Each selected connection is represented by two directed edges for message passing, with $E_{op}=E_{po}^{\top}$. Given state $s$, $\pi_e$ predicts the connection probability of each candidate node pair $(u,v)$ from their concatenated features. Here, $\mathbf h_i$ denotes the corresponding row of $\mathbf X$ for an original node $i\in\mathcal V$, or of $\mathbf P$ for a prompt node $i\in\mathcal U$:
\begin{equation}
\mathbf{z}_{uv} = [\mathbf{h}_u ; \mathbf{h}_v] \in \mathbb{R}^{2d}.
\label{policy-concat}
\end{equation}

An MLP $g_e$ followed by a sigmoid activation produces the connection probability:
\begin{equation}
p_{uv} = \sigma\!\left(g_e(\mathbf{z}_{uv})\right), \quad p_{uv} \in (0,1).
\label{policy-prob}
\end{equation}

The binary connection decision is then sampled as: 
\begin{equation}
e_{uv} \sim \mathrm{Bernoulli}(p_{uv}),
\label{policy-sample}
\end{equation}
where $e_{uv}=1$ adds an edge between $u$ and $v$, and $e_{uv}=0$ leaves them unconnected.

\noindent\textbf{Connection Relative Policy Optimization.} We seek to update the edge-level policy $\pi_e$ through within-group reward comparisons, drawing on GRPO's relative advantage estimation~\citep{GRPO}. However, the frozen GNN produces deterministic predictions for a fixed input rather than sampled responses for group comparison. We therefore sample complete connection patterns at its input under a shared prompt-node scheme and develop Connection Relative Policy Optimization (CRPO) to learn from the relative task rewards of the resulting augmented graphs. Let $\mathcal C$ denote the set of candidate undirected prompt--original and prompt--prompt connections allowed by the current prompt-node scheme. Each complete action $a_k^e$ specifies all binary connection decisions over $\mathcal C$. For each group, we fix the input graph, prompt count, and prompt features, and sample $K$ complete actions under the same state $s$ using the old policy $\pi_e^{\mathrm{old}}$ at the start of sampling. Each action produces a prompt-augmented graph $\mathcal G^{\prime(k)}$.

The pre-trained GNN and task head remain frozen throughout Stage~I. Each augmented graph is evaluated using the available labels from the training split to obtain the task loss $\mathcal L_{\mathrm{task}}^{(k)}$, whose negative value serves as the reward. The relative advantage of the $k$-th candidate connection pattern is computed as:
\begin{equation}
    A_{\mathrm{rel}}^{(k)}
    =
    \frac{-\mathcal{L}_{\mathrm{task}}^{(k)}
    -\mu_{\mathrm{group}}}
    {\sigma_{\mathrm{group}}+\varepsilon_{\mathrm{num}}},
    \label{CRPO_A}
\end{equation}

where $\mu_{\mathrm{group}}$ and $\sigma_{\mathrm{group}}$ are the mean and standard deviation of the rewards for the $K$ candidate connection patterns of the same input, and $\varepsilon_{\mathrm{num}}>0$ is a numerical stability constant. Only the connection pattern varies within a group, allowing complete connection actions to be evaluated through their relative task performance.

Given a state, candidate edges are sampled independently from the Bernoulli distributions defined above. The probability of a complete connection action is:
\begin{equation}
\pi_e(a_k^e\mid s)=\prod_{(u,v)\in\mathcal C}p_{uv}^{e_{uv}^{(k)}}(1-p_{uv})^{1-e_{uv}^{(k)}},
\label{crpo-joint-probability}
\end{equation}
where $e_{uv}^{(k)}$ is the binary decision for candidate connection $(u,v)$ in the $k$-th complete action, and $p_{uv}$ is its connection probability under the corresponding policy. During policy updates, we retain the sampled states and actions, their old-policy log-probabilities, and the group-relative advantages. The likelihood ratio compares the current and old policies on the same sampled action:
\begin{equation}
\rho^{(k)}=\frac{\pi_e(a_k^e\mid s)}{\pi_e^{\mathrm{old}}(a_k^e\mid s)},
\label{crpo-likelihood-ratio}
\end{equation}
where the numerator is the probability of the sampled action under the current policy, and the denominator is its old-policy probability recorded during sampling. The candidate edge set and prompt-node scheme remain fixed throughout the policy updates for this group. The clipped surrogate objective is:

\begin{multline}
\mathcal{L}_{\mathrm{clip}} = -\mathbb{E}_{k\in K} \biggl[
\min \Bigl( \rho^{(k)}A_{\mathrm{rel}}^{(k)}, \\
\mathrm{clip} \bigl( \rho^{(k)}, 1-\epsilon, 1+\epsilon \bigr)
A_{\mathrm{rel}}^{(k)} \Bigr) \biggr], \hspace{-\rightmargin}
\label{L1_GRPO}
\end{multline}

where $\epsilon$ controls the clipping range of the likelihood ratio. Policy entropy regularization maintains exploration over connection patterns:
\begin{equation}
{
\mathcal{L}_{\mathrm{entropy}}^{\mathrm{I}}
=
\mathbb{E}\left[H(\pi_e(\cdot|s))\right],
}
\label{L2_GRPO}
\end{equation}

where $H(\pi_e(\cdot\mid s))$ is the entropy of the complete connection-action distribution. The entropy term and clipped objective form the Stage~I training objective:
\begin{equation}
{
\mathcal{L}_{\mathrm{total}}^{\mathrm{I}}
=
\mathcal{L}_{\mathrm{clip}}
-
\beta\mathcal{L}_{\mathrm{entropy}}^{\mathrm{I}},
}
\label{loss_CRPO}
\end{equation}
where $\beta$ controls the strength of entropy regularization. Before each gradient update, we compute the batch-averaged KL divergence $D_{\mathrm{KL}}(\pi_e^{\mathrm{old}}\|\pi_e)$. If it exceeds a threshold $\delta$, we stop further CRPO updates on the current batch of sampled data.

\subsection{Stage II: Node-level Policy Learning}
\label{stage2}

\noindent\textbf{Motivation 2.} The limited range of prompt-node counts makes direct configuration prediction a coarse formulation for learning a policy to refine prompt-node configurations. Downstream evaluation provides only a holistic performance signal for each resulting prompt-node scheme, offering little guidance on how the current configuration should be adjusted for the training inputs. We therefore formulate prompt-node configuration as an incremental evolution process. Each decision adjusts the prompt count of the current prompt-augmented graph and generates prompt features consistent with the resulting state. The state transitions induced by these hybrid actions define an MDP, transforming one-shot configuration selection into a sequence of state-dependent decisions. This formulation provides richer local performance comparisons and allows the prompt-count and feature-generation policies to be jointly learned along state trajectories through proximal policy optimization.

\noindent\textbf{Node-level Policy.} The node-level policy comprises a prompt-count policy $\pi_n$ and a feature-generation policy $\pi_p$. Stage~II starts from the prompt configuration of the current alternating cycle. At step $t$, let $M_t$ and $\mathbf P_t$ denote the prompt count and features, respectively, and let $\mathcal G'_t$ denote the corresponding augmented graph. The state $s_t$ contains the current shared prompt count and a batch-level representation. For each augmented graph, we first mean-pool the node representations produced by the frozen encoder:
\begin{equation}
\bar{\mathbf H}_t=\frac{1}{N+M_t}\sum_{i=1}^{N+M_t}\mathbf h_{i,t},
\end{equation}
where $\mathbf h_{i,t}$ is the encoded representation of node $i$ in the current augmented graph. For a training batch $\mathcal B$, the node policy uses the batch mean $\widetilde{\mathbf H}_t=|\mathcal B|^{-1}\sum_{\mathcal G\in\mathcal B}\bar{\mathbf H}_{\mathcal G,t}$ and generates one prompt-node scheme shared by its inputs. Here, $\bar{\mathbf H}_{\mathcal G,t}$ denotes the pooled representation above for graph $\mathcal G$. The count network $g_n$ maps the pooled representation to two logits. A softmax yields the probabilities of adding or removing a prompt node, from which the count action is sampled:
\begin{equation}
a_t^n\sim\pi_n(\cdot\mid s_t)=\operatorname{softmax}(g_n(\widetilde{\mathbf H}_t)),
\end{equation}
where addition and removal correspond to $\Delta_t=+1$ and $\Delta_t=-1$, respectively. The updated prompt count is constrained to the prescribed range:
\begin{equation}
k_t=M_{t+1}=\operatorname{clip}(M_t+\Delta_t,1,M_{\max}),
\end{equation}
where $M_{\max}$ is the maximum prompt count. At a boundary, an action that would exceed the allowed range leaves the count unchanged.

After the count action determines $k_t$, the feature-generation policy $\pi_p$ produces a feature action $a_t^p\in\mathbb R^{k_t\times d}$. Two feature networks, $g_\mu$ and $g_\sigma$, map the same state representation to mean and standard deviation matrices, respectively:
\begin{equation}
\boldsymbol\mu_t=g_\mu(\widetilde{\mathbf H}_t),\quad
\boldsymbol\sigma_t=\exp(g_\sigma(\widetilde{\mathbf H}_t)),
\end{equation}
where both network outputs are reshaped into $M_{\max}\times d$ matrices indexed by prompt position, $d$ is the prompt feature dimension, and the elementwise exponential ensures positive standard deviations. During training, feature components are sampled independently, and the first $k_t$ rows are retained according to the updated prompt count:
\begin{equation}
Z_{t,j\ell}\sim\mathcal N(\mu_{t,j\ell},\sigma_{t,j\ell}^2),\quad
a_t^p=\mathbf P_{t+1}=\mathbf Z_t[1\!:\!k_t,:],
\label{feature-sampling}
\end{equation}
where $j=1,\ldots,M_{\max}$ and $\ell=1,\ldots,d$. Let $a_t=(a_t^n,a_t^p)$, and let $\omega$ collect the parameters of the count network and the two feature networks. The joint node-action distribution is:
\begin{equation}
\pi_\omega(a_t\mid s_t)=\pi_n(a_t^n\mid s_t)\pi_p(a_t^p\mid s_t,k_t),
\label{node-joint-policy}
\end{equation}

\begin{algorithm}[!t]
\caption{TTPR}
\label{alg:TTPR}
\footnotesize
\begin{algorithmic}[1]
\STATE \textbf{Input:} $f_\theta,\mathcal D_L,\mathcal D_V,K,L,M_{\max},\gamma,\lambda,\beta,\delta,\epsilon,T$
\STATE Initialize $\pi_e,\pi_n,\pi_p,V_\nu,g_\psi$ and freeze $f_\theta$
\STATE $M\leftarrow\lceil M_{\max}/2\rceil,\quad P_{j\ell}\overset{\mathrm{iid}}{\sim}\mathcal N(0,1),\quad q^*\leftarrow-\infty$
\FOR{epoch $i=1,\ldots,T$}
    \STATE \textbf{Stage I: CRPO} (update $\pi_e$ only)
    \FOR{each training batch $\mathcal B\subseteq\mathcal D_L$}
        \STATE $\pi_e^{\mathrm{old}}\leftarrow\pi_e,\quad\mathcal B_e\leftarrow[\,]$
        \STATE $(M,\mathbf P,\_)\leftarrow\operatorname{Rollout}_L(\mathcal B,M,\mathbf P\mid\pi_n,\pi_p,\pi_e^{\mathrm{old}})$
        \FOR{each $(\mathcal G,y)\in\mathcal B$}
            \STATE Form $s$ from $(\mathcal G,M,\mathbf P)$
            \STATE $a_k^e\sim\pi_e^{\mathrm{old}}(\cdot\mid s),\quad k=1,\ldots,K$
            \STATE Construct $\mathcal G^{\prime(k)}$,\quad $r^{(k)}\leftarrow-\mathcal L_{\mathrm{task}}(\mathcal G^{\prime(k)},y)$
            \STATE $A_{\mathrm{rel}}^{(k)}\leftarrow$ Eq.~\eqref{CRPO_A}
            \STATE Cache $(s,a_k^e,\log\pi_e^{\mathrm{old}}(a_k^e\mid s),A_{\mathrm{rel}}^{(k)})$ in $\mathcal B_e$
        \ENDFOR
        \STATE Update $\pi_e$ on $\mathcal B_e$ by Eq.~\eqref{loss_CRPO} with KL limit $\delta$
    \ENDFOR
    \STATE \textbf{Stage II: INPPO} (update $\omega,\nu$ only)
    \FOR{each training batch $\mathcal B\subseteq\mathcal D_L$}
        \STATE $(\pi_{\omega^{\mathrm{old}}},V_{\mathrm{old}})\leftarrow(\pi_\omega,V_\nu),\quad\mathcal B_n\leftarrow[\,]$
        \STATE $(M_0,\mathbf P_0)\leftarrow(M,\mathbf P)$
        \STATE Construct $\{\mathcal G'_0\}_{\mathcal G\in\mathcal B}$ with $\pi_e$ and compute $\widetilde{\mathbf H}_0$ using $f_\theta$
        \FOR{$t=0,\ldots,L-1$}
            \STATE $s_t\leftarrow(M_t,\widetilde{\mathbf H}_t),\quad a_t^n\sim\pi_n^{\mathrm{old}}(\cdot\mid s_t)$
            \STATE $M_{t+1}\leftarrow\operatorname{clip}(M_t+\Delta_t,1,M_{\max})$
            \STATE $\mathbf P_{t+1}\sim\pi_p^{\mathrm{old}}(\cdot\mid s_t,M_{t+1})$ [Eq.~\eqref{feature-sampling}]
            \STATE Construct $\{\mathcal G'_{t+1}\}_{\mathcal G\in\mathcal B}$ with fixed $\pi_e$
            \STATE Compute $\widetilde{\mathbf H}_{t+1}$ using $f_\theta$,\quad $s_{t+1}\leftarrow(M_{t+1},\widetilde{\mathbf H}_{t+1})$
            \STATE $r_t\leftarrow-\frac{1}{|\mathcal B|}\sum_{(\mathcal G,y)\in\mathcal B}\mathcal L_{\mathrm{task}}(\mathcal G'_{t+1},y)$
            \STATE Cache $(s_t,(a_t^n,\mathbf P_{t+1}),r_t,s_{t+1})$ in $\mathcal B_n$
            \STATE Cache $\{\bar{\mathbf H}_{\mathcal G,t}\}_{\mathcal G\in\mathcal B}$ and old log probabilities/values in $\mathcal B_n$
        \ENDFOR
        \STATE $(\hat A_t,\hat R_t)\leftarrow$ GAE [Eqs.~\eqref{td-error}--\eqref{gae-targets}]
        \STATE $A_t^\lambda\leftarrow\operatorname{Normalize}_{\mathcal B_n}(\hat A_t)$
        \STATE Update $(\omega,\nu)$ on $\mathcal B_n$ by Eq.~\eqref{loss_INPPO} with KL limit $\delta$
        \STATE $(M,\mathbf P)\leftarrow(M_L,\mathbf P_L)$
    \ENDFOR
    \STATE \textbf{Task-head update} (update $\psi$ only)
    \FOR{each training batch $\mathcal B\subseteq\mathcal D_L$}
        \STATE $M_0\leftarrow\lceil M_{\max}/2\rceil,\quad P_{0,j\ell}\overset{\mathrm{iid}}{\sim}\mathcal N(0,1)$
        \STATE $(M_L,\mathbf P_L,\mathcal B'_L)\leftarrow\operatorname{Rollout}_L(\mathcal B,M_0,\mathbf P_0\mid\pi_n,\pi_p,\pi_e)$
        \STATE Update $\psi$ with $\frac{1}{|\mathcal B|}\sum_{(\mathcal G'_L,y)\in\mathcal B'_L}\mathcal L_{\mathrm{task}}(\mathcal G'_L,y)$
    \ENDFOR
    \STATE $M_0\leftarrow\lceil M_{\max}/2\rceil,\quad P_{0,j\ell}\overset{\mathrm{iid}}{\sim}\mathcal N(0,1)$
    \STATE $(M^*,\mathbf P^*,\_)\leftarrow\operatorname{Rollout}_L(\mathcal D_L,M_0,\mathbf P_0\mid\pi_n,\pi_p,\pi_e)$
    \STATE $q\leftarrow\operatorname{Evaluate}(\mathcal D_V\mid\pi_e,g_\psi,M^*,\mathbf P^*)$
    \STATE If $q>q^*$, save $C^*\leftarrow(\pi_e,\pi_n,\pi_p,g_\psi,M^*,\mathbf P^*)$ and $q^*\leftarrow q$
\ENDFOR
\STATE \textbf{Output:} $C^*$
\end{algorithmic}
\end{algorithm}
where $\pi_p$ is the product of the Gaussian component densities over the first $k_t$ rows. The edge-level policy $\pi_e$, with its parameters fixed, constructs connections from the updated prompt count and features to form $\mathcal G'_{t+1}$. The frozen encoder then computes the next state $s_{t+1}$. Each count adjustment builds on the preceding configuration, and the feature distribution is conditioned on the current augmented-graph representation, passing state information from the preceding configuration to subsequent decisions.

\noindent\textbf{Incremental Node Proximal Policy Optimization.} We propose Incremental Node Proximal Policy Optimization (INPPO), which builds on PPO's clipped policy objective~\citep{PPO} to jointly learn count adjustments and conditional feature generation from task feedback along incremental configuration trajectories. Before collecting trajectories, we save the old policy $\pi_{\omega^{\mathrm{old}}}$ and keep it fixed during collection and updates on the current batch. At each step, the count action and prompt features are sampled from the old policy. The resulting augmented graphs are evaluated to obtain the batch-averaged task reward:
\begin{equation}
r_t=-\frac{1}{|\mathcal B|}\sum_{(\mathcal G,y)\in\mathcal B}\mathcal L_{\mathrm{task}}(\mathcal G'_{t+1},y),
\end{equation}
 where $\mathcal G'_{t+1}$ is the augmented graph corresponding to input $\mathcal G$, and $y$ denotes its available training labels. The parameters of the edge-level policy, pre-trained encoder, and task head remain fixed throughout Stage~II. Executing $L$ steps yields transitions $(s_t,a_t,r_t,s_{t+1})$, together with the recorded old-policy log probability of each action.

The value network is implemented as an MLP $g_v$ parameterized by $\nu$:
\begin{equation}
V_\nu(s_t)=\frac{1}{|\mathcal B|}\sum_{\mathcal G\in\mathcal B}g_v(\bar{\mathbf H}_{\mathcal G,t})\in\mathbb R,
\end{equation}
where the value estimates recorded during collection are denoted by $V_{\mathrm{old}}$. Task rewards and value estimates are used to compute generalized advantage estimates and value regression targets:
\begin{equation}
\delta_t=r_t+\gamma(1-d_t)V_{\mathrm{old}}(s_{t+1})-V_{\mathrm{old}}(s_t),
\label{td-error}
\end{equation}
\begin{equation}
\hat A_t=\delta_t+\gamma\lambda(1-d_t)\hat A_{t+1},\quad
\hat R_t=\hat A_t+V_{\mathrm{old}}(s_t),
\label{gae-targets}
\end{equation}
where $\gamma$ is the discount factor, $\lambda$ controls advantage estimation, and $d_t$ indicates trajectory termination. The complete process of generating a configuration terminates after $L$ steps, with $d_{L-1}=1$ and $d_t=0$ at earlier steps. The recursion starts at $\hat A_L=0$ and uses a zero next-state value at the final step. Value targets are constructed from the unnormalized $\hat A_t$. Advantages are normalized within the current batch to obtain $A_t^\lambda$ for policy updates.

During updates, the current joint probability is recomputed for the same sampled count action and feature matrix. Let $\ell_t(\omega)=\log\pi_\omega(a_t\mid s_t)$ denote the joint log probability. The likelihood ratio between the current and old policies is:
\begin{equation}
\rho_t(\omega)=\exp\!\left(\ell_t(\omega)-\ell_t(\omega^{\mathrm{old}})\right),
\end{equation}
where the feature log probability is summed over the retained rows and feature dimensions. The clipped policy loss for INPPO is:
\begin{equation}
\mathcal L_{\mathrm{INPPO}}
=-\mathbb E_t[\min\{\rho_t A_t^\lambda,
\operatorname{clip}(\rho_t,1-\epsilon,1+\epsilon)A_t^\lambda\}],
\end{equation}
where $\epsilon$ controls the clipping range of the joint likelihood ratio. Old log probabilities, advantages, and value regression targets remain fixed during updates on the current batch. The value network is trained to regress the trajectory targets:
\begin{equation}
\mathcal L_{\mathrm{VF}}(\nu)=\mathbb E_t[(V_\nu(s_t)-\hat R_t)^2],
\end{equation}
where the expectation is taken over the current batch of sampled data. Entropy regularization uses the entropy of the joint node-action distribution:
\begin{equation}
\mathcal L_{\mathrm{entropy}}^{\mathrm{II}}
=\mathbb E_t[H(\pi_\omega(\cdot\mid s_t))],
\end{equation}
where the joint entropy consists of the count-action entropy and the conditional feature entropy. The conditional feature entropy is summed over the valid components for each count $k_t$ and then averaged under the current count policy. The overall Stage~II objective is:
\begin{equation}
\mathcal L_{\mathrm{total}}^{\mathrm{II}}
=\mathcal L_{\mathrm{INPPO}}+\mathcal L_{\mathrm{VF}}
-\beta\mathcal L_{\mathrm{entropy}}^{\mathrm{II}},
\label{loss_INPPO}
\end{equation}
where $\beta$ controls the strength of entropy regularization. The joint policy loss updates the count network and both feature networks that generate the means and standard deviations.

Before each gradient update, we compute the mean KL divergence from the old joint policy to the current joint policy over states sampled in the current batch. This divergence consists of the KL divergence between the count distributions and that between the conditional feature distributions. The latter is summed over the valid feature dimensions and weighted by the old count policy. If the batch-averaged KL divergence exceeds $\delta$, further INPPO updates on the current batch of sampled data are stopped. Together with the edge-level policy learned in Stage~I, the learned node-level policy constructs final prompt configurations.

\begin{table}[tbp]
\renewcommand{\arraystretch}{1.05}
\centering
\caption{Statistics of datasets for node classification.}
\resizebox{1.0\linewidth}{!}{
\begin{tabular}{lrrrrrrr}
\toprule
Dataset & Nodes & Edges & Classes & MinN & MaxN & Hops \\
\midrule
Cora       & 2708   &  5429   & 7  & 5 & 150 & 4 \\
CiteSeer   & 3327   &  9104   & 6  & 20 & 150 & 3 \\
PubMed     & 19717  & 88648   & 3  & 50 & 300 & 2 \\
Computers  & 13752  & 491722  & 10 & 100 & 1000 & 2 \\
Photo      & 7650   & 238162  & 8  & 200 & 2000 & 2 \\
\bottomrule
\end{tabular}
}
\label{tab:node_dataset_stats}
\end{table}

\begin{table}[tbp]
\renewcommand{\arraystretch}{1.05}
\centering
\caption{Statistics of datasets for graph classification.}
\fontsize{8pt}{9.6pt}\selectfont
\setlength{\tabcolsep}{3pt}
\begin{tabular*}{\linewidth}{@{\extracolsep{\fill}}c|cc|c|cc@{}}
\toprule
\textbf{Dataset} & \textbf{Molecules} & \textbf{Tasks} & \textbf{Dataset} & \textbf{Molecules} & \textbf{Tasks} \\
\cmidrule{1-6}
BBBP & 2,039 & 1 &Tox21 & 7,831 & 12 \\

ClinTox & 1,478 & 2  & HIV & 41,127 & 1  \\

BACE & 1,513 & 1& MUV & 93,087 & 17\\

ToxCast & 8,575 & 617 & SIDER & 1,427 & 27 \\
\bottomrule
\end{tabular*}
\label{tab:stat_graph}
\end{table}

\begin{table}[tbp]
\renewcommand{\arraystretch}{1.05}
\centering
\caption{Key hyperparameters and typical candidate values used in our experiments.}
\resizebox{0.99\linewidth}{!}{
\begin{tabular}{lll}
\toprule
\textbf{Category} & \textbf{Parameter} & \textbf{Candidate Values} \\
\midrule
\multirow{7}{*}{Architecture} 
 &  GNN layers (node / graph) &  2 / 5 \\
 & GNN type                & \{GIN\} \\
 &  Node feature dim after SVD & \{100\} \\
 &  Molecular atom embedding dim & \{300\} \\
 &  GNN hidden dim (node / graph) &  128 / 300 \\
 & Task head layers & 2 \\
 & Task hidden dim         & \{128,256\} \\
\midrule
\multirow{5}{*}{Task/Training} 
 & Epochs                  & \{50\} \\
 & Train batch size        & \{8,16,32\} \\
 & Minibatch size          & \{5,10,15,20\} \\
 & Learning rate of task optimization      & \{1e-2,5e-3\} \\
 & Gamma (discount factor) & \{0.99\} \\
\midrule
\multirow{11}{*}{RL-specific} 

 & Group size $K$     & \{2,4,8,16,32,64\} \\
 & Trajectory length $L$ & \{5,10,15,20,25,30\} \\

 & Max prompt nodes $M_{\max}$ & \{3,5,7,9,11\} \\
 &Learning rate of Policy $\pi_e$          & \{5e-4,1e-3,5e-3\} \\
 &Learning rate of Policy $\pi_n$           & \{5e-4,1e-3,5e-3\} \\
 &Learning rate of Policy $\pi_p$         & \{5e-4,1e-3,5e-3\} \\
 &Learning rate of Value Function $V$                  & \{5e-4,1e-3,5e-3\} \\
 & Clip param $\epsilon$             & \{0.2\} \\
 & GAE coefficient $\lambda$         & \{0.90,0.95,1.00\} \\
 & Entropy coefficient $\beta$    & \{0.005,0.01,0.02,0.05\} \\
 & KL threshold $\delta$    & \{0.01\} \\

\bottomrule
\end{tabular}
}
\label{tab:impl_params}
\end{table}

\subsection{Task Head Optimization}

After the two policy-learning stages, we freeze the encoder $f_\theta$ and the policy parameters.  For each training batch, we initialize $M_0=\lceil M_{\max}/2\rceil$ and independently sample the entries of $\mathbf P_0$ from a standard normal distribution. The edge-level policy constructs the initial augmented graph. At each step, the node-level policy adjusts the prompt count and generates features, and the edge-level policy samples connections to update the graph state. After $L$ steps, we encode the final graph $\mathcal G'_L$ and feed its mean-pooled representation to the task head $g_\psi$. We compute $\mathcal L_{\mathrm{task}}$ using the available training labels and update only $\psi$. Edge-level policy learning, node-level policy learning, and task-head updates alternate for $T$ cycles. Model parameters are retained across cycles, and the updated task head provides feedback for the next policy-learning cycle.

After task-head optimization in each epoch, a rollout over the training set produces a shared prompt count and feature matrix, which are retained with the checkpoint selected by validation performance. At inference, all model parameters and this prompt-node scheme remain fixed. The edge policy constructs connections separately for each input, and $g_\psi$ predicts from the resulting augmented-graph representation. The settings of $L$ and $M_{\max}$ are given in Table~\ref{tab:impl_params}.

\begin{table*}[tbp]
\renewcommand{\arraystretch}{1.05}

\centering
\small
\caption{
Accuracy (\%) and Macro-F1 (\%) (with standard deviations) of various prompt-tuning strategies in the 10-shot node classification task.
\textsuperscript{\dag} indicates results cited directly
from RELIEF~\citep{RELIEF}, \textsuperscript{\ddag} indicates results cited directly from LEAP~\citep{LEAP}, and \textsuperscript{$\ast$} represents results reproduced using official code.
\textbf{Boldface} marks the highest performance among all models and \underline{underlining} denotes the second-best performance.}
\label{node-results}

\setlength{\tabcolsep}{4pt}
\resizebox{1.0\textwidth}{!}{
\begin{tabular}{@{}l|cccccccccc@{}}
\toprule
\multirow{2}{*}{\textbf{Method}} &
\multicolumn{2}{c}{\textbf{Cora}} &
\multicolumn{2}{c}{\textbf{CiteSeer}} &
\multicolumn{2}{c}{\textbf{PubMed}} &
\multicolumn{2}{c}{\textbf{Computers}} &
\multicolumn{2}{c}{\textbf{Photo}} \\
\cmidrule(lr){2-3} \cmidrule(lr){4-5} \cmidrule(lr){6-7} \cmidrule(lr){8-9} \cmidrule(lr){10-11}
 & \textbf{Acc} & \textbf{F1} & \textbf{Acc} & \textbf{F1} & \textbf{Acc} & \textbf{F1} & \textbf{Acc} & \textbf{F1} & \textbf{Acc} & \textbf{F1} \\
\midrule

\rowcolor{gray!15} \multicolumn{11}{@{}l}{\textbf{Pre-training Strategy: GPPT}} \\ 
FT\textsuperscript{\dag} & 54.40  $\pm$  1.44 & 53.27  $\pm$  1.45 & 61.28  $\pm$  0.93 & 56.60  $\pm$  0.92 & 71.40  $\pm$  1.61 & 67.11  $\pm$  1.51 & \underline{80.52  $\pm$  1.18} & 76.35  $\pm$  1.57 & 83.40  $\pm$  3.00 & 81.31  $\pm$  2.88 \\ 
GPPT\textsuperscript{\dag} & 45.76  $\pm$  0.34 & 46.05  $\pm$  0.30 & 57.00  $\pm$  0.55 & 51.03  $\pm$  0.68 & 71.36  $\pm$  0.70 & 64.53  $\pm$  0.84 & 75.80  $\pm$  0.42 & 69.97  $\pm$  0.57 & 82.48  $\pm$  0.74 & 80.65  $\pm$  0.71\\ 

GPF\textsuperscript{\dag} & 55.04  $\pm$  1.82 & 54.20  $\pm$  2.03 & 60.48  $\pm$  1.11 & 55.57  $\pm$  0.97 & 64.96  $\pm$  6.77 & 60.63  $\pm$  5.17 & 73.32  $\pm$  2.56 & 68.53  $\pm$  2.21 & 83.08  $\pm$  0.68 & 80.60  $\pm$  1.04 \\ 
GPF-plus\textsuperscript{\dag} & 56.36  $\pm$  1.46 & 55.84  $\pm$  1.33 & 61.36  $\pm$  0.70 & 56.50  $\pm$  0.44 & 72.32  $\pm$  4.17 & \underline{67.52  $\pm$  3.51} & \textbf{80.56  $\pm$  1.97} & \underline{76.69  $\pm$  1.87} & 85.08  $\pm$  1.22 & 82.26  $\pm$  0.94 \\
SUPT$_{\text{soft}}$\textsuperscript{\dag} & 55.24  $\pm$  1.01 & 54.69  $\pm$  1.18 & 59.60  $\pm$  1.97 & 54.88  $\pm$  1.45 & 62.40  $\pm$  7.67 & 58.76  $\pm$  6.17 & 77.28  $\pm$  4.18 & 72.63  $\pm$  4.86 & 84.52  $\pm$  1.55 & 82.22  $\pm$  1.36 \\ 

SUPT$_{\text{hard}}$\textsuperscript{\dag} & 56.96  $\pm$  1.28 & 55.86  $\pm$  1.28 & 61.28  $\pm$  1.84 & 56.81  $\pm$  1.45 & 71.64  $\pm$  5.98 & 65.20  $\pm$  5.70 & 77.96  $\pm$  1.61 & 73.60  $\pm$  1.72 & \underline{85.24  $\pm$  0.81} & 82.92  $\pm$  0.56 \\

PSP\textsuperscript{$\ast$} & 55.48 $\pm$ 1.42 & 54.93 $\pm$ 1.08 & 59.44 $\pm$ 0.89 & 54.72 $\pm$ 1.32 & 62.76 $\pm$ 4.26 & 59.12 $\pm$ 4.17 & 77.08 $\pm$ 5.10 & 72.43 $\pm$ 3.82 & 84.64 $\pm$ 1.95 & 82.34 $\pm$ 1.07 \\
All-in-One\textsuperscript{$\ast$} & 54.24  $\pm$  1.91 & 52.10  $\pm$  1.86 & 63.20  $\pm$  1.72& 57.43  $\pm$  1.75 & 67.56  $\pm$  2.52 & 66.80  $\pm$  1.95 & 74.12  $\pm$  1.29 & 73.50  $\pm$  1.45 & 81.84  $\pm$  0.94 & 79.53  $\pm$  0.71\\ 

EdgePrompt\textsuperscript{$\ast$} & 58.20  $\pm$  2.21 & 52.18  $\pm$  3.14 
& 59.52  $\pm$  2.83 & 55.59  $\pm$  2.56 
& 62.92  $\pm$  3.19 & 58.88  $\pm$  2.88
& 72.16  $\pm$  1.95 & 69.60  $\pm$  1.10 
& 80.12  $\pm$  0.72 & 79.55  $\pm$  0.94 \\ 

EdgePrompt$_+$\textsuperscript{$\ast$} & \underline{62.96  $\pm$  2.35} & \underline{61.22  $\pm$  2.75} & 63.76  $\pm$  3.32 & 56.40  $\pm$  2.97 & 65.24  $\pm$  2.59 & 59.80  $\pm$  3.40 & 73.24  $\pm$  1.76 & 71.80  $\pm$  1.12 & 81.84  $\pm$  1.80 & 81.20  $\pm$  0.86 \\ 

RELIEF\textsuperscript{\dag} & 56.76  $\pm$  0.23 & 55.94  $\pm$  0.25 & \underline{65.16  $\pm$  1.21} & \underline{58.55  $\pm$  1.12} & \underline{72.36  $\pm$  1.05} & 66.71  $\pm$  1.10 & 77.68  $\pm$  0.93 & 72.53  $\pm$  1.26 & 85.16  $\pm$  0.89 & \underline{84.07  $\pm$  0.69} \\ 
LEAP\textsuperscript{\ddag} & 58.32  $\pm$  1.50 & 56.81  $\pm$  0.94 & 64.24  $\pm$  1.56 & 57.67  $\pm$  1.54 & 71.68  $\pm$  1.48 & 66.65  $\pm$  1.61 & 75.48  $\pm$  1.33 & 71.09  $\pm$  1.09 & 85.20  $\pm$  1.50 & 83.35  $\pm$  1.42 \\
\cmidrule{1-11}
TTPR & \textbf{65.16  $\pm$  0.64} & \textbf{63.21  $\pm$  1.25} & \textbf{73.00  $\pm$  1.24} & \textbf{69.95  $\pm$  0.37} & \textbf{77.12  $\pm$  0.43} & \textbf{75.12  $\pm$  0.59} & 79.04  $\pm$  0.34 & \textbf{78.15  $\pm$  0.94} & \textbf{86.52  $\pm$  0.10} & \textbf{85.16  $\pm$  0.89} \\

\cmidrule{1-11}

\rowcolor{gray!15} \multicolumn{11}{@{}l}{\textbf{Pre-training Strategy: GPrompt}} \\ 
FT\textsuperscript{\dag} & 60.60  $\pm$  1.89 & 59.82  $\pm$  2.01 & 60.88  $\pm$  2.79 & 56.90  $\pm$  2.20 & 72.20  $\pm$  1.02 & 67.44  $\pm$  0.81 & 80.64  $\pm$  1.23 & 76.02  $\pm$  1.75 & 82.24  $\pm$  1.92 & 79.71  $\pm$  1.79 \\
GPrompt\textsuperscript{\dag} & 57.20  $\pm$  2.13 & 53.98  $\pm$  2.82 & 59.52  $\pm$  0.65 & 55.46  $\pm$  0.70 & 70.00  $\pm$  0.62 & 66.17  $\pm$  0.51 & 67.44  $\pm$  0.80 & 64.81  $\pm$  1.42 & 74.08  $\pm$  2.55 & 71.26  $\pm$  2.66 \\

GPF\textsuperscript{\dag} &  59.12  $\pm$  1.95 & 58.70  $\pm$  1.92 & 63.24  $\pm$  3.09 & 55.00  $\pm$  2.54 & 71.24  $\pm$  1.98 & 66.35  $\pm$  2.84 & 80.68  $\pm$  1.48 & 76.29  $\pm$  1.60 & 83.64  $\pm$  1.30 & 81.33  $\pm$  1.27 \\ 
GPF-plus\textsuperscript{\dag} & 60.60  $\pm$  1.08 & 60.63  $\pm$  0.87 & 62.44  $\pm$  1.86 & 56.81  $\pm$  0.55 & 73.64  $\pm$  3.73 & 69.95  $\pm$  3.67 & 80.68  $\pm$  0.65 & 76.93  $\pm$  0.92 & 86.20  $\pm$  1.15 & 83.57  $\pm$  1.19 \\
SUPT$_{\text{soft}}$\textsuperscript{\dag} &  60.28  $\pm$  1.32 & 60.05  $\pm$  1.08 & 63.16  $\pm$  2.35 & 55.95  $\pm$  1.69 & 73.52  $\pm$  3.81 & 69.70  $\pm$  3.31 & 80.84  $\pm$  0.85 & 75.30  $\pm$  0.81 & 85.36  $\pm$  1.75 & 82.47  $\pm$  1.94 \\
SUPT$_{\text{hard}}$\textsuperscript{\dag} & 58.92  $\pm$  0.86 & 58.75  $\pm$  0.82 & 64.32  $\pm$  2.38 & 55.96  $\pm$  2.20 & 73.56  $\pm$  1.97 & 69.48  $\pm$  2.18 & 80.52  $\pm$  0.95 & 76.69  $\pm$  0.87 & 86.28  $\pm$  0.55 & 83.92  $\pm$  0.34 \\

PSP\textsuperscript{$\ast$} & 60.08 $\pm$ 1.11 & 59.85 $\pm$ 0.68 & 63.44 $\pm$ 1.97 & 56.23 $\pm$ 1.42 & 73.20 $\pm$ 3.98 & 69.38 $\pm$ 2.42 & 79.64 $\pm$ 1.22 & 75.14 $\pm$ 0.71 & 85.60 $\pm$ 1.44 & 82.71 $\pm$ 1.80 \\
All-in-One\textsuperscript{$\ast$} & 58.80  $\pm$  0.82 & 57.12  $\pm$  1.15 & 65.44  $\pm$  0.95 & 56.12  $\pm$  1.24 & 70.96  $\pm$  1.64 & 67.58  $\pm$  1.55 & 79.80  $\pm$  0.47 & 76.60  $\pm$  0.92 & 85.52  $\pm$  0.82 & \underline{84.15  $\pm$  0.74} \\ 

EdgePrompt\textsuperscript{$\ast$} 
& 62.36  $\pm$  2.26& 61.44  $\pm$  2.95 &
61.80  $\pm$  1.81 & 55.66  $\pm$  2.56 & 70.60  $\pm$  3.93 &68.12  $\pm$  3.67 & 
77.40  $\pm$  1.10 & 74.80  $\pm$  1.00 & 83.80  $\pm$  0.80 & 81.42  $\pm$  0.57 \\ 
EdgePrompt$_+$\textsuperscript{$\ast$} & \underline{63.12  $\pm$  0.97} & \underline{62.30  $\pm$  2.90} & 62.60  $\pm$  2.20 & 57.30  $\pm$  2.64 & 71.64  $\pm$  4.50 & 68.80  $\pm$  4.56 & 79.80  $\pm$  1.92 & 76.95  $\pm$  1.46 & 84.12  $\pm$  0.98 & 83.44  $\pm$  0.75 \\ 

RELIEF\textsuperscript{\dag} & 61.20  $\pm$  1.20 & 60.89  $\pm$  0.91 & \underline{66.28  $\pm$  0.27} & \underline{58.01  $\pm$  0.32} & \underline{74.96  $\pm$  1.72} & \underline{70.22  $\pm$  1.70} & \textbf{82.04  $\pm$  0.83} & 78.03  $\pm$  0.77 & \underline{86.48  $\pm$  0.90} & 83.88  $\pm$  0.62 \\
LEAP\textsuperscript{\ddag} & 61.32  $\pm$  0.92 & 61.12  $\pm$  1.14 & 65.92  $\pm$  1.01 & 57.90  $\pm$  0.85 & 71.84  $\pm$  1.14 & 66.95  $\pm$  1.21 & \underline{81.32  $\pm$  1.55} & \underline{78.21  $\pm$  1.60} & 86.40  $\pm$  1.39 & 83.95  $\pm$  1.32 \\

\cmidrule{1-11}
TTPR & \textbf{67.92  $\pm$  2.12} & \textbf{66.84  $\pm$  1.70} & \textbf{72.36  $\pm$  1.36} & \textbf{69.92  $\pm$  1.03} & \textbf{76.52  $\pm$  0.88} & \textbf{74.41  $\pm$  0.82} & 81.16  $\pm$  0.34 & \textbf{78.65  $\pm$  0.94} & \textbf{87.24  $\pm$  0.27} & \textbf{85.72  $\pm$  1.04} \\ 

\bottomrule
\end{tabular}
}
\end{table*}

\section{Experiments}

\noindent{\textbf{Datasets and Baselines:} We evaluate node classification on Cora, CiteSeer, and PubMed~\citep{citation-dataset}, and Computers and Photo~\citep{purchase-dataset}. For graph classification, we use BBBP, ClinTox, BACE, Tox21, ToxCast, HIV, MUV, and SIDER from MoleculeNet~\citep{MoleculeNet}. We compare TTPR with FT, GPPT~\citep{GPPT}, GPrompt~\citep{GPrompt}, GPF and GPF-Plus~\citep{GPF}, SUPT$_{\mathrm{soft}}$ and SUPT$_{\mathrm{hard}}$~\citep{SUPT}, EdgePrompt and EdgePrompt$_+$~\citep{EdgePrompt}, All-in-One~\citep{ALL-in-One}, PSP~\citep{PSP2024}, RELIEF~\citep{RELIEF}, and LEAP~\citep{LEAP}.}

\noindent\textbf{Experimental Setup:}
For backbone pre-training, we adopt GPPT~\citep{GPPT} and GPrompt~\citep{GPrompt} for node classification, AttrMasking~\citep{masking-contextpred}, ContextPred~\citep{masking-contextpred}, Infomax~\citep{infomax}, GCL~\citep{GraphCL} for graph classification.  Reproduced baselines use the same backbone, preprocessing, and data partitions as TTPR. Results taken from prior studies are identified in the tables.
 For node classification, we randomly sample 10 labeled nodes per class for training and select two disjoint sets of 500 remaining nodes for validation and testing, respectively. All experiments are performed on subgraphs sampled from the original dataset graphs. For each target node, we construct its induced subgraph by expanding to a fixed hop distance and enforcing upper and lower bounds on the number of nodes retained (MaxN, MinN, and Hops), with settings reported in Section~\ref{sec:experimental-details}.
 For graph classification, we partition each dataset into training, validation, and test sets using an 80/10/10 scaffold split. We then randomly sample a total of 50 graphs from the training partition for few-shot adaptation, while retaining the validation and test partitions for evaluation. Further implementation details are provided in Section~\ref{sec:experimental-details}.

\begin{table*}[tbp]
\renewcommand{\arraystretch}{1.07}

\centering
\caption{ROC-AUC (\%) with standard deviations for prompt-tuning methods under four pre-training strategies on the 50-shot graph classification task. \textsuperscript{\dag} indicates results cited directly from RELIEF~\citep{RELIEF}, \textsuperscript{\ddag} indicates results cited directly from LEAP~\citep{LEAP}, and \textsuperscript{$\ast$} represents results reproduced using official code. \textbf{Boldface} marks the highest performance among all models and \underline{underlining} denotes the second-best performance. Avg. is the arithmetic mean across the eight datasets.}
\label{tab:main_graph_results}
\setlength{\tabcolsep}{8pt}
\resizebox{\textwidth}{!}{
\begin{tabular}{@{} l | c c c c c c c c | c @{}}
\toprule
\textbf{Tuning Strategy} &
\textbf{BBBP} &
\textbf{Tox21} &
\textbf{ToxCast} &
\textbf{SIDER} &
\textbf{ClinTox} &
\textbf{MUV} &
\textbf{HIV} &
\textbf{BACE} &
\textbf{Avg.} \\
\midrule

\rowcolor{gray!15} \multicolumn{10}{@{}l}{\textbf{Pre-training Strategy: AttrMasking}} \\

FT\textsuperscript{\dag} & 66.48  $\pm$  0.44 & 72.32  $\pm$  0.19 & 57.35  $\pm$  0.42 & 54.62  $\pm$  0.58 & 59.71  $\pm$  4.43 & 61.49  $\pm$  1.04 & 58.37  $\pm$  1.51 & 65.41  $\pm$  2.09 & 61.97 \\
GPF\textsuperscript{\dag} & 66.67  $\pm$  0.60 & 72.31  $\pm$  0.30 & 58.01  $\pm$  0.27 & 55.65  $\pm$  1.92 & 67.25  $\pm$  2.41 & 62.58  $\pm$  0.31 & 58.71  $\pm$  2.56 & 68.54  $\pm$  1.24 & 63.72 \\
GPF-plus\textsuperscript{\dag} & 66.29  $\pm$  0.36 & 72.75  $\pm$  0.38 & 57.91  $\pm$  0.38 & 55.05  $\pm$  1.20 & 68.89  $\pm$  2.86 & 62.70  $\pm$  0.48 & 58.71  $\pm$  1.07 & 68.07  $\pm$  1.20 & 63.80 \\
SUPT$_{\text{soft}}$\textsuperscript{\dag} & 66.28  $\pm$  0.81 & \underline{72.76  $\pm$  0.41} & 58.28  $\pm$  0.34 & 54.70  $\pm$  1.03 & 68.58  $\pm$  4.54 & 62.81  $\pm$  0.25 & 58.95  $\pm$  1.71 & 66.41  $\pm$  0.49 & 63.60 \\
SUPT$_{\text{hard}}$\textsuperscript{\dag} & 66.93  $\pm$  0.91 & 72.75  $\pm$  0.41 & 58.18  $\pm$  0.44 & 55.20  $\pm$  1.26 & 68.36  $\pm$  3.57 & 62.81  $\pm$  0.22 & 59.02  $\pm$  1.61 & 68.92  $\pm$  0.96 & 64.02 \\
PSP\textsuperscript{$\ast$} & 66.29 $\pm$ 0.86 & 72.75 $\pm$ 0.41 & 58.23 $\pm$ 0.41 & 55.13 $\pm$ 1.12 & 68.74 $\pm$ 3.22 & 62.76 $\pm$ 0.37 & 58.99 $\pm$ 1.38 & 67.38 $\pm$ 1.01 & 63.78 \\
All-in-One\textsuperscript{$\ast$} & 65.70  $\pm$  1.03 & 69.41  $\pm$  0.43 & 58.33  $\pm$  0.54 & 55.90  $\pm$  0.44 & 70.74  $\pm$  2.59 & 60.30  $\pm$  1.53 & 59.12  $\pm$  1.14 & 66.69  $\pm$  1.05 & 63.27 \\
EdgePrompt\textsuperscript{$\ast$} & 66.22  $\pm$  0.75 & 70.20  $\pm$  0.38 & 56.41  $\pm$  0.79 & 53.40  $\pm$  0.32 & 64.13  $\pm$  2.23 & 60.80  $\pm$  1.12 & 57.22 $\pm$  1.44 & 65.42  $\pm$  1.15 & 61.73 \\
EdgePrompt$_+$\textsuperscript{$\ast$} & 66.35  $\pm$  0.47 & 71.11  $\pm$  0.55 & 57.12  $\pm$  0.56 & 55.12  $\pm$  0.43 & 68.20  $\pm$  2.54 & 60.92  $\pm$  1.03 & 58.01  $\pm$  1.32 & 67.16  $\pm$  0.94 & 63.00 \\
RELIEF\textsuperscript{\dag} & \underline{67.18  $\pm$  0.55} & 72.40  $\pm$  0.32 & \underline{58.41  $\pm$  0.14} & \underline{56.54  $\pm$  0.75} & \textbf{74.39  $\pm$  1.18} & 63.18  $\pm$  0.17 & \underline{59.42  $\pm$  0.55} & 69.37  $\pm$  0.70 & \underline{65.11} \\
LEAP\textsuperscript{\ddag} & 66.95  $\pm$  0.35 & \textbf{72.79  $\pm$  0.22} & \textbf{58.43  $\pm$  0.20} & 56.17  $\pm$  0.91 & 73.33  $\pm$  2.22 & \underline{63.42  $\pm$  0.20} & 59.33  $\pm$  0.82 & \underline{69.45  $\pm$  0.88} & 64.98 \\
\cmidrule{1-10}
TTPR & \textbf{68.29  $\pm$  0.56} & 71.84  $\pm$  0.32 & 58.31  $\pm$  0.25 & \textbf{58.12  $\pm$  0.63} & \underline{73.92  $\pm$  3.25} & \textbf{63.77  $\pm$  0.59} & \textbf{61.33  $\pm$  0.78} & \textbf{72.21  $\pm$  0.14} & \textbf{65.97} \\
\cmidrule{1-10}

\rowcolor{gray!15} \multicolumn{10}{@{}l}{\textbf{Pre-training Strategy: ContextPred}} \\

FT\textsuperscript{\dag} & 62.82  $\pm$  0.83 & 70.11  $\pm$  0.38 & 57.68  $\pm$  0.69 & 56.68  $\pm$  0.78 & 60.99  $\pm$  2.28 & 61.28  $\pm$  1.93 & 59.14  $\pm$  1.93 & 64.47  $\pm$  2.47 & 61.65 \\
GPF\textsuperscript{\dag} & 61.65  $\pm$  0.76 & 70.42  $\pm$  0.29 & 58.51  $\pm$  0.38 & 56.55  $\pm$  0.46 & 58.80  $\pm$  2.57 & 64.83  $\pm$  0.32 & 59.68  $\pm$  2.20 & 68.57  $\pm$  1.19 & 62.38 \\
GPF-plus\textsuperscript{\dag} & 61.25  $\pm$  0.73 & 70.12  $\pm$  0.42 & 57.64  $\pm$  0.70 & 56.86  $\pm$  0.52 & 61.43  $\pm$  3.06 & 64.59  $\pm$  0.28 & 59.15  $\pm$  2.23 & 68.58  $\pm$  1.42 & 62.45 \\
SUPT$_{\text{soft}}$\textsuperscript{\dag} & 61.85  $\pm$  1.56 & 70.37  $\pm$  0.14 & 57.82  $\pm$  0.37 & 56.57  $\pm$  0.50 & 59.13  $\pm$  1.99 & 64.59  $\pm$  1.00 & 56.58  $\pm$  1.94 & 68.92  $\pm$  0.92 & 61.98 \\
SUPT$_{\text{hard}}$\textsuperscript{\dag} & 62.29  $\pm$  1.41 & 70.40  $\pm$  0.15 & 58.06  $\pm$  0.39 & 56.34  $\pm$  0.72 & 59.18  $\pm$  2.07 & 64.57  $\pm$  0.69 & 59.08  $\pm$  1.47 & 69.40  $\pm$  1.62 & 62.42 \\
PSP\textsuperscript{$\ast$} & 61.55 $\pm$ 1.14 & 70.25 $\pm$ 0.29 & 57.81 $\pm$ 0.53 & 56.51 $\pm$ 0.58 & 59.69 $\pm$ 2.40 & 64.59 $\pm$ 0.85 & 57.90 $\pm$ 1.71 & 68.75 $\pm$ 1.30 & 62.13 \\
All-in-One\textsuperscript{$\ast$} & 61.22  $\pm$  0.86 & 69.90  $\pm$  0.69 & 57.80  $\pm$  0.66 & 56.44  $\pm$  0.63 & 60.20  $\pm$  2.72 & 64.62  $\pm$  1.14 & 56.72  $\pm$  1.48 & 67.19  $\pm$  1.17 & 61.76 \\
EdgePrompt\textsuperscript{$\ast$} & 61.75  $\pm$  0.75 & 69.44  $\pm$  0.45 & 56.32  $\pm$  0.68 & 55.32  $\pm$  0.41 & 59.72  $\pm$  1.72 & 62.32  $\pm$  1.25 & 55.20  $\pm$  1.12 & 67.62  $\pm$  1.19 & 60.96 \\
EdgePrompt$_+$\textsuperscript{$\ast$} & 62.12  $\pm$  0.64 & 70.12  $\pm$  0.52 & 58.01  $\pm$  0.97 & 56.11  $\pm$  0.52 & 59.98  $\pm$  2.03 & 62.55  $\pm$  1.53 & 56.34  $\pm$  1.44 & 68.35  $\pm$  1.08 & 61.70 \\
RELIEF\textsuperscript{\dag} & \underline{62.99  $\pm$  0.67} & \underline{70.51  $\pm$  0.14} & \textbf{58.58  $\pm$  0.04} & 56.89  $\pm$  0.25 & \underline{63.06  $\pm$  1.74} & \underline{64.87  $\pm$  0.31} & 59.72  $\pm$  1.67 & 70.19  $\pm$  0.96 & \underline{63.35} \\
LEAP\textsuperscript{\ddag} & 62.98  $\pm$  0.81 & \textbf{70.68  $\pm$  0.17} & \underline{58.53  $\pm$  0.35} & \textbf{56.96  $\pm$  0.58} & 62.29  $\pm$  1.65 & \textbf{64.92  $\pm$  0.61} & \underline{59.99  $\pm$  1.58} & \underline{70.33  $\pm$  0.84} & 63.34 \\
\cmidrule{1-10}
TTPR & \textbf{63.48  $\pm$  0.25} & 70.39  $\pm$  0.24 & 57.42  $\pm$  0.28 & \underline{56.95  $\pm$  0.70} & \textbf{64.13  $\pm$  2.19} & 63.72  $\pm$ 0.70 & \textbf{61.89  $\pm$  1.99} & \textbf{73.01  $\pm$  1.16} & \textbf{63.87} \\
\cmidrule{1-10}

\rowcolor{gray!15} \multicolumn{10}{@{}l}{\textbf{Pre-training Strategy: Infomax}} \\

FT\textsuperscript{$\dag$} & 65.26  $\pm$  1.05 & 71.54  $\pm$  0.73 & 57.98  $\pm$  0.42 & 53.45  $\pm$  0.49 & 59.01  $\pm$  2.18 & 64.08  $\pm$  0.84 & 65.13  $\pm$  2.50 & 64.57  $\pm$  0.71 & 62.63 \\
GPF\textsuperscript{$\dag$} & 66.30  $\pm$  0.94 & 71.26  $\pm$  0.30 & 58.33  $\pm$  0.27 & 53.65  $\pm$  0.34 & 63.11  $\pm$  2.35 & 67.19  $\pm$  0.10 & 64.09  $\pm$  0.66 & 64.32  $\pm$  0.88 & 63.53 \\
GPF-plus\textsuperscript{$\dag$} & 66.12  $\pm$  1.27 & 71.54  $\pm$  0.63 & 58.43  $\pm$  0.27 & 53.76  $\pm$  1.15 & 65.06  $\pm$  1.56 & 67.20  $\pm$  0.17 & 64.81  $\pm$  1.62 & 65.88  $\pm$  1.72 & 64.10 \\
SUPT$_{\text{soft}}$\textsuperscript{$\dag$} & 66.45  $\pm$  0.85 & \underline{71.82  $\pm$  0.16} & 58.71  $\pm$  0.44 & 53.72  $\pm$  0.45 & 65.96  $\pm$  3.37 & 66.52  $\pm$  0.55 & 65.51  $\pm$  2.31 & 66.26  $\pm$  2.45 & 64.37 \\
SUPT$_{\text{hard}}$\textsuperscript{$\dag$} & 66.14  $\pm$  0.85 & 71.55  $\pm$  0.37 & 58.65  $\pm$  0.33 & 53.82  $\pm$  0.49 & 65.90  $\pm$  3.41 & 66.53  $\pm$  0.35 & 65.59  $\pm$  2.26 & 66.40  $\pm$  0.70 & 64.32 \\
PSP\textsuperscript{$\ast$} & 66.13 $\pm$ 0.94 & 71.55 $\pm$ 0.40 & 58.54 $\pm$ 0.39 & 53.79 $\pm$ 0.47 & 65.93 $\pm$ 2.98 & 66.53 $\pm$ 0.45 & 65.16 $\pm$ 1.94 & 66.33 $\pm$ 1.39 & 64.25 \\
All-in-One\textsuperscript{$\ast$} & 65.95  $\pm$  1.03 & 70.21  $\pm$  0.43 & 57.32  $\pm$  0.54 & \underline{55.15  $\pm$  0.44} & \underline{67.71  $\pm$  2.59} & 65.30  $\pm$  1.53 & 64.21  $\pm$  1.14 & 66.65  $\pm$  1.05 & 64.06 \\
EdgePrompt\textsuperscript{$\ast$} & 65.91  $\pm$  0.85 & 69.91  $\pm$  0.59 & 56.25  $\pm$  0.51 & 53.04  $\pm$  0.42 & 63.20  $\pm$  2.73 & 64.80  $\pm$  1.53 & 63.79  $\pm$  1.56 & 66.19  $\pm$  0.94 & 62.89 \\
EdgePrompt$_+$\textsuperscript{$\ast$} & 67.12  $\pm$  0.95 & 70.90  $\pm$  0.38 & 58.12  $\pm$  0.41 & 53.42  $\pm$  0.66 & 64.33  $\pm$  2.34 & 65.12  $\pm$  1.61 & 64.12  $\pm$  1.08 & \underline{68.21  $\pm$  0.64} & 63.92 \\
RELIEF\textsuperscript{$\dag$} & 67.93  $\pm$  0.73 & 71.58  $\pm$  0.13 & \underline{58.78  $\pm$  0.17} & 53.95  $\pm$  0.40 & 66.28  $\pm$  2.07 & \underline{67.23  $\pm$  0.08} & \underline{67.00  $\pm$  0.55} & 67.95  $\pm$  0.52 & 65.09 \\
LEAP\textsuperscript{\ddag} & \underline{68.27  $\pm$  1.09} & \textbf{71.85  $\pm$  0.21} & \textbf{58.79  $\pm$  0.22} & 54.00  $\pm$  0.56 & 66.44  $\pm$  2.29 & \textbf{67.28  $\pm$  0.49} & \textbf{67.15  $\pm$  1.38} & 67.31  $\pm$  0.91 & \underline{65.14} \\
\cmidrule{1-10}
TTPR & \textbf{68.65  $\pm$  0.92} & 71.32  $\pm$  0.77 & 58.41  $\pm$  0.50 & \textbf{56.57  $\pm$  0.24 } & \textbf{69.96  $\pm$  2.72} & 66.64  $\pm$  2.69 & 65.84  $\pm$  1.23 & \textbf{72.91  $\pm$  0.75} & \textbf{66.29 } \\
\cmidrule{1-10}

\rowcolor{gray!15} \multicolumn{10}{@{}l}{\textbf{Pre-training Strategy: GCL}} \\

FT\textsuperscript{$\dag$} & 62.13  $\pm$  1.66 & 61.35  $\pm$  0.88 & 53.96  $\pm$  0.80 & 52.63  $\pm$  0.71 & 68.73  $\pm$  4.74 & 51.96  $\pm$  1.93 & 58.67  $\pm$  1.41 & 49.94  $\pm$  3.80 & 57.42 \\
GPF\textsuperscript{$\dag$} & 61.58  $\pm$  1.81 & 59.92  $\pm$  1.29 & 54.44  $\pm$  0.31 & 51.21  $\pm$  0.56 & 75.52  $\pm$  3.08 & 52.26  $\pm$  1.28 & 58.37  $\pm$ 1.16 & 56.09 $\pm$ 2.44 & 58.67 \\
GPF-plus\textsuperscript{$\dag$} & 62.19  $\pm$  1.45 & 60.13  $\pm$  0.52 & 54.43  $\pm$  0.43 & 50.90  $\pm$  0.78 & 75.50  $\pm$  1.11 & 52.40  $\pm$  1.55 & 58.30  $\pm$  1.68 & 59.55 $\pm$ 3.17 & 59.18 \\
SUPT$_{\text{soft}}$\textsuperscript{$\dag$} & 63.96  $\pm$  0.85 & 60.13  $\pm$  0.52 & 54.57  $\pm$  0.76 & 51.44  $\pm$  1.24 & 78.03  $\pm$  2.32 & 51.86  $\pm$  1.48 & 59.30  $\pm$  2.59 & 57.39  $\pm$  2.64 & 59.59 \\
SUPT$_{\text{hard}}$\textsuperscript{$\dag$} & 64.15  $\pm$  0.96 & 60.56  $\pm$  0.20 & 54.72  $\pm$  0.71 & 51.60  $\pm$  0.85 & 77.88  $\pm$  2.06 & 51.78  $\pm$  1.72 & 59.24  $\pm$  2.42 & 59.68  $\pm$  1.46 & 59.95 \\
PSP\textsuperscript{$\ast$} & 63.43 $\pm$ 1.21 & 60.35 $\pm$ 0.52 & 54.65 $\pm$ 0.74 & 51.52 $\pm$ 0.85 & 76.69 $\pm$ 2.19 & 52.13 $\pm$ 1.64 & 59.27 $\pm$ 2.51 & 59.56 $\pm$ 2.91 & 59.70 \\
All-in-One\textsuperscript{$\dag$} & 62.90  $\pm$  3.77 & \underline{61.49  $\pm$  0.96} & 54.72  $\pm$ 1.03 & 52.73  $\pm$  0.84 & 66.59  $\pm$  3.84 & 53.34  $\pm$  4.20 & 60.60  $\pm$  3.80 & 59.57  $\pm$  5.62 & 58.99 \\
EdgePrompt\textsuperscript{$\ast$} & 61.25  $\pm$  0.72 & 59.42  $\pm$  0.53 & 53.12  $\pm$  0.34 & 52.54  $\pm$  0.62 & 69.55  $\pm$  1.97 & 50.80  $\pm$  1.13 & 59.72  $\pm$  1.25 & 56.98  $\pm$  0.98 & 57.92 \\
EdgePrompt$_+$\textsuperscript{$\ast$} & 64.44  $\pm$  0.85 & 60.19  $\pm$  0.59 & 54.66  $\pm$  0.51 & \underline{54.40  $\pm$  0.42} & 74.20  $\pm$  2.73 & 52.72  $\pm$  1.53 & \underline{60.89  $\pm$  1.56} & 60.12  $\pm$  0.94 & 60.20 \\
RELIEF\textsuperscript{$\dag$} & \underline{64.80  $\pm$  0.69} & 61.45 $\pm$  0.35 & \underline{55.03  $\pm$  0.39} & 52.82  $\pm$  0.59 & \underline{79.75  $\pm$  1.28} & 53.52  $\pm$  2.17 & 60.44  $\pm$  2.39 & \underline{62.97  $\pm$  1.19} & \underline{61.35} \\
LEAP\textsuperscript{\ddag} & 63.55  $\pm$  1.08 & \textbf{61.54  $\pm$  0.39} & 54.89  $\pm$  0.35 & 52.33  $\pm$  1.01 & \textbf{79.93  $\pm$  1.96} & \underline{53.59  $\pm$  1.91} & 60.53  $\pm$  1.98 & 62.50  $\pm$  1.41 & 61.11 \\
\cmidrule{1-10}
TTPR & \textbf{64.92  $\pm$  1.23} & 60.68  $\pm$  0.26 & \textbf{55.51  $\pm$  0.99} & \textbf{55.29  $\pm$  1.42} & 77.54  $\pm$  1.98 & \textbf{57.21  $\pm$  0.61} & \textbf{62.28  $\pm$  0.80} & \textbf{70.36  $\pm$  0.64} & \textbf{62.97} \\
\bottomrule
\end{tabular}
}
\label{tab:appendix_graph_results}
\end{table*}

\subsection{Experimental Details}\label{sec:experimental-details}

\textbf{Dataset Details.} We train and evaluate TTPR on both node classification and graph classification tasks. Cora, CiteSeer, and PubMed are citation networks, while Computers and Photo are co-purchase graphs. For graph classification, we adopt eight molecular property prediction benchmarks from MoleculeNet. Each molecule is represented as an attributed graph, where nodes correspond to atoms and edges correspond to chemical bonds, and labels indicate biomedical endpoints. These benchmarks span a wide range of scales, from 1.4k to 93k molecules, and cover both single-task and multi-task classification with 1--617 prediction targets. We evaluate node classification using Accuracy and Macro-F1, and graph classification using ROC-AUC.

\noindent\textbf{Implementation Details.}  Experiments are conducted on an NVIDIA GeForce RTX 4090 GPU. For node classification, input features are reduced to 100 dimensions using SVD before pre-training and prompt tuning. We train for a fixed number of epochs and select hyperparameters and checkpoints based on validation performance. Task-specific architecture settings and training hyperparameters are summarized in Table~\ref{tab:impl_params}.

\begin{table*}[tbp]
\renewcommand{\arraystretch}{1.05}
\centering
\caption{Ablation results (Acc, Macro-F1, and ROC-AUC in \%). The reporting protocol is mean $\pm$ standard deviation over five independent runs. Node classification uses GPPT and graph classification uses AttrMasking.}
\label{tab:ablation_results}
\label{tab:twostage_vs_singlestage}
\fontsize{7.25pt}{8.70pt}\selectfont
\setlength{\tabcolsep}{0pt}
\begin{tikzpicture}
\node[inner sep=0pt, outer sep=0pt] (ablationtable) {%
\begin{tabular}{@{}p{0.1\textwidth}*{9}{>{\centering\arraybackslash}p{0.1\textwidth}}@{}}
\toprule
 & \multicolumn{6}{c}{Node Classification} & \multicolumn{3}{c}{Graph Classification} \\[-1.2pt]
\cmidrule(lr){2-7}\cmidrule(lr){8-10}
\noalign{\vskip -1.2pt}
Variant & \multicolumn{2}{c}{CiteSeer} & \multicolumn{2}{c}{PubMed} & \multicolumn{2}{c}{Photo} & BACE & BBBP & HIV \\[-1.2pt]
\cmidrule(lr){2-3}\cmidrule(lr){4-5}\cmidrule(lr){6-7}
\noalign{\vskip -1.2pt}
 & Acc & F1 & Acc & F1 & Acc & F1 & ROC-AUC & ROC-AUC & ROC-AUC \\[-0.5pt]
\midrule
TTPR & \textbf{73.00 $\pm$ 1.24} & \textbf{69.95 $\pm$ 0.37} & \textbf{77.12 $\pm$ 0.43} & \textbf{75.12 $\pm$ 0.59} & \textbf{86.52 $\pm$ 0.10} & \textbf{85.16 $\pm$ 0.89} & \textbf{72.21 $\pm$ 0.14} & \textbf{68.29 $\pm$ 0.56} & \textbf{61.33 $\pm$ 0.78} \\
\midrule
w/o INPPO & 69.36 $\pm$ 1.46 & 65.82 $\pm$ 1.21 & 70.96 $\pm$ 1.17 & 68.23 $\pm$ 1.36 & 82.36 $\pm$ 0.74 & 80.53 $\pm$ 1.13 & 68.14 $\pm$ 0.62 & 65.64 $\pm$ 1.28 & 59.46 $\pm$ 1.07 \\
w/o CRPO & 68.72 $\pm$ 1.18 & 64.93 $\pm$ 1.57 & 74.84 $\pm$ 0.84 & 72.36 $\pm$ 0.92 & 82.44 $\pm$ 1.26 & 80.68 $\pm$ 1.55 & 66.58 $\pm$ 0.83 & 66.04 $\pm$ 1.09 & 57.76 $\pm$ 1.34 \\
w/o Count & 71.84 $\pm$ 0.98 & 68.17 $\pm$ 0.83 & 76.56 $\pm$ 0.61 & 74.43 $\pm$ 0.78 & 83.64 $\pm$ 0.73 & 81.92 $\pm$ 1.01 & 71.63 $\pm$ 0.38 & 67.12 $\pm$ 0.97 & 60.87 $\pm$ 0.93 \\
w/o Feature & 70.92 $\pm$ 1.07 & 67.48 $\pm$ 1.04 & 72.64 $\pm$ 1.09 & 70.08 $\pm$ 1.18 & 84.84 $\pm$ 0.48 & 83.23 $\pm$ 0.82 & 69.36 $\pm$ 0.54 & 66.28 $\pm$ 1.16 & 60.42 $\pm$ 0.88 \\
w/o N-RL & 71.24 $\pm$ 1.09 & 67.68 $\pm$ 0.72 & 73.44 $\pm$ 1.08 & 71.61 $\pm$ 1.15 & 85.76 $\pm$ 0.39 & 84.62 $\pm$ 0.76 & 68.87 $\pm$ 0.67 & 67.84 $\pm$ 0.91 & 58.63 $\pm$ 1.12 \\
\midrule
w/o Alt & 68.24 $\pm$ 1.42 & 64.68 $\pm$ 1.31 & 71.12 $\pm$ 0.86 & 70.90 $\pm$ 1.10 & 78.96 $\pm$ 1.13 & 77.31 $\pm$ 1.49 & 67.81 $\pm$ 0.21 & 64.34 $\pm$ 0.64 & 58.10 $\pm$ 1.20 \\

w/o Alt-EN & 67.92 $\pm$ 0.92 & 64.12 $\pm$ 1.11 & 71.08 $\pm$ 0.95 & 70.22 $\pm$ 0.83 & 81.36 $\pm$ 1.24 & 78.31 $\pm$ 1.02 & 68.21 $\pm$ 0.21 & 65.12 $\pm$ 0.33 & 59.05 $\pm$ 0.90  \\

w/o Alt-NE & 70.32 $\pm$ 1.42 & 66.72 $\pm$ 0.79 & 73.04 $\pm$ 0.69 & 71.66 $\pm$ 0.95 & 80.12 $\pm$ 1.22 & 79.21 $\pm$ 1.12 & 69.11 $\pm$ 0.41 & 65.34 $\pm$ 0.60 & 59.55 $\pm$ 0.97 \\
\bottomrule
\end{tabular}};
\draw[line width=0.4pt, dash pattern=on 2pt off 2pt]
  ([xshift=0.7\textwidth]ablationtable.north west) --
  ([xshift=0.7\textwidth]ablationtable.south west);
\end{tikzpicture}

\par

\end{table*}

\subsection{Main Results}

As shown in Table~\ref{node-results}, TTPR achieves the highest accuracy in eight of the ten node classification settings and the highest Macro-F1 in all ten settings. The gains are more pronounced on citation networks. On CiteSeer, TTPR improves Macro-F1 over the strongest baseline by 11.40 and 11.91 percentage points under GPPT and GPrompt pre-training, respectively. Sparse citation networks provide limited local context, and learned prompt features and connections introduce additional task information that helps the frozen encoder form representations suited to downstream classification. Denser graphs such as Computers already contain rich neighborhood relations, leaving less room for improvements in overall accuracy. On Computers, TTPR achieves the highest Macro-F1 under both pre-training strategies despite lower accuracy than the best baseline. Macro-F1 weights all classes equally, highlighting improvements in average per-class performance that are less evident from overall accuracy.

We further evaluate TTPR on 50-shot molecular graph classification under four pre-training strategies. Table~\ref{tab:main_graph_results} shows that TTPR achieves the highest average ROC-AUC under each strategy. This consistent advantage suggests that task-specific propagation and aggregation can complement knowledge learned under different pre-training objectives. Prompt nodes participate in both message passing and graph readout, providing additional paths to combine information beyond the original bond neighborhoods. Learning their features and connectivity allows the graph representation to adapt to the downstream objective while preserving the original molecular bonds and keeping the encoder frozen.

{
\begin{table}[t]
\fontsize{8pt}{9.6pt}\selectfont
\setlength{\tabcolsep}{3pt}

\makeatletter\def\@captype{figure}\makeatother
\centering
\includegraphics[width=1.0\linewidth]{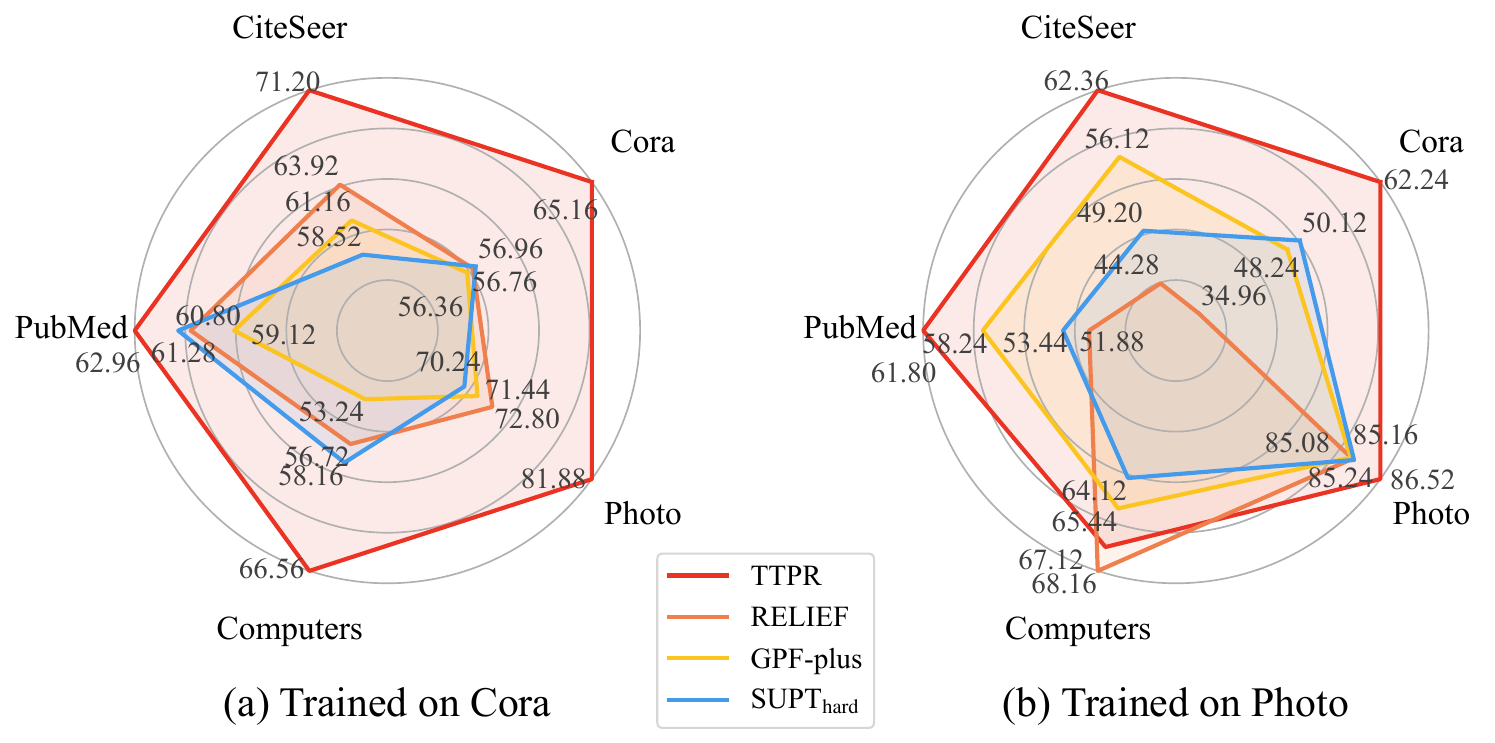}
\caption{Cross-dataset accuracy (\%) under GPPT pre-training when transferring from (a) Cora and (b) Photo. TTPR transfers frozen source policies to each target dataset, using its own frozen pretrained GNN and a classifier trained on its 10-shot training set. Results are averaged over five independent runs with different random seeds.}

\label{fig:cross_dataset}

\end{table}
\begin{table}[t]
\fontsize{8pt}{9.6pt}\selectfont
\setlength{\tabcolsep}{3pt}
\begingroup
\makeatletter\def\@captype{table}\makeatother
\renewcommand{\arraystretch}{1.05}

\centering
\caption{Average training time per epoch across different methods. TTPR includes CRPO, INPPO, and task-head training.}
\fontsize{7.25pt}{8.70pt}\selectfont
\begin{tabular*}{0.99\linewidth}{@{\extracolsep{\fill}}lrrr|rrr@{}}
\toprule
Method & Cora & PubMed & Photo & BACE & BBBP & MUV \\
\midrule
GPF         & 7.9s  & 8.5s  & 19.2s & 10.3s  & 15.1s  & 22.6s \\
SUPT$_{\mathrm{hard}}$ & 8.8s  & 9.9s  & 20.4s  & 10.9s  & 16.0s  & 24.3s \\
All-in-One  & 8.2s  & 10.7s  & 19.7s  & 12.2s  & 17.9s  & 30.2s \\

RELIEF      & 37.5s & 81.4s & 168.1s & 55.7s & 53.8s& 503.2s  \\
LEAP        & 33.4s & 47.0s & 98.4s & 39.3s & 49.2s & 279.5s \\
TTPR        & 31.2s & 37.1s & 62.5s  & 78.3s  & 51.2s & 114.6s \\
\bottomrule
\end{tabular*}
\label{tab:times}

\endgroup
\end{table}

\subsection{Ablation Study}

We evaluate prompt configuration components and training schemes on CiteSeer, PubMed, and Photo node classification with GPPT pre-training, and BACE, BBBP, and HIV graph classification with AttrMasking pre-training. w/o INPPO fixes both the prompt count and randomly initialized features to assess the node-learning module as a whole. w/o CRPO replaces the learned connection policy with a fixed cosine-threshold rule to compare learned and heuristic connections. w/o Count assesses adaptive prompt capacity by fixing the count while retaining feature learning, whereas w/o Feature fixes the randomly initialized features, uses the rows corresponding to the current count, and retains count learning to examine the contribution of learned prompt content. Fixed counts are selected from $\{1,3,5,7,9,11\}$ within the corresponding full model's count limit, and thresholds from $\{0.1,0.3,0.5,0.7,0.9\}$. Each candidate is trained separately, with the best fixed value selected on the validation set. Each variant retains the remaining components and their update rules. 

To compare incremental RL with direct task-loss training, w/o N-RL replaces INPPO with direct count prediction and feature learning while retaining CRPO and alternating training. It predicts count probabilities in one pass and jointly learns the count predictor and prompt features using the probability-weighted task losses of candidate counts, treating the generated connections as fixed during these updates. During inference, it selects the most probable count. This comparison changes both the incremental decision process and the optimization method. {To assess whether repeated mutual adaptation helps coordinate prompt content and connection structure, we compare three training variants. w/o Alt jointly optimizes count, feature, and connection decisions as a single action using PPO. w/o Alt-EN trains CRPO followed by INPPO, holding the other component fixed in each phase, while w/o Alt-NE reverses this order. Both sequential variants start from the same initialization as TTPR and exhaust the first phase's training budget before entering the second, without returning to the first. Both variants retain the original optimizers under matched training budgets and an identical task-head update protocol.}

Table~\ref{tab:ablation_results} shows that removing either INPPO or CRPO reduces average performance. Fixed thresholds rely on feature similarity, whereas CRPO evaluates connection patterns through task feedback. Its gains may therefore reflect the importance of propagating prompt content through connections suited to the task. PubMed is more sensitive to fixed features, while Photo is more sensitive to a fixed count. Counts constrain prompt capacity and the space of possible connections, while features determine the injected information. These distinct roles may explain their varying contributions across datasets. Retaining either learning component outperforms fixing both, and joint learning achieves higher average performance. Compared with direct prediction in w/o N-RL, INPPO can progressively refine configurations using the updated augmented-graph state. TTPR outperforms all three training variants across the reported metrics, suggesting that coordinated node and edge learning benefits from both separate updates and continued mutual adaptation. The degradation of w/o Alt, which places all decisions under a single PPO objective, is consistent with the difficulty of distinguishing node and edge contributions in joint feedback. The sequential variants retain separate optimization, but the component trained first cannot continue adapting to subsequent configuration changes. The generally stronger performance of w/o Alt-NE suggests that learning connections after prompt content is established may be more effective, whereas changing prompt features under a frozen edge policy may weaken their learned compatibility. Both sequential variants outperform w/o Alt on graph classification, while w/o Alt-EN offers no corresponding gains on CiteSeer or PubMed, indicating that the benefits of separate updates depend on both training order and dataset. Alternating training combines conditional updates with repeated mutual adaptation, supporting more effective joint configurations.

\begin{table*}[tbp]
\makeatletter\def\@captype{figure}\makeatother
    \centering
    \includegraphics[width=0.96\textwidth]{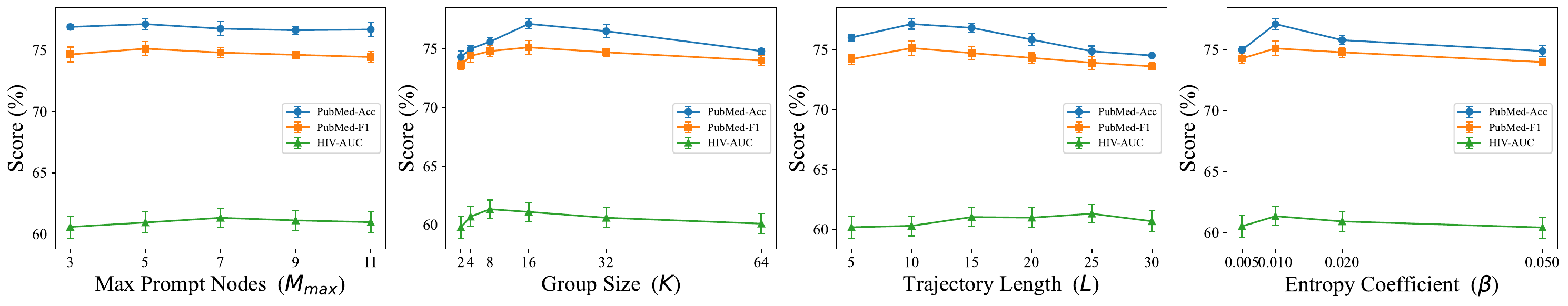}
    \caption{Parameter sensitivity analysis evaluating the impact of maximum prompt nodes ($M_{max}$), group size ($K$), trajectory length ($L$), and entropy coefficient ($\beta$) on the PubMed and HIV datasets.}
    \label{fig:param_analysis}
\end{table*}

\subsection{Cross-Dataset Generalization}

We evaluate the cross-dataset generalization of prompt configuration policies by learning them on Cora or Photo and transferring them to multiple target datasets. Node features are independently reduced to 100 dimensions using SVD for each dataset, which uses a GNN pretrained on its own data with GPPT. All encoders have the same architecture and representation dimensionality, and their parameters remain frozen during policy learning and target-domain adaptation. For each target dataset, we transfer and freeze the source-trained edge policy, prompt-count policy, and prompt feature generator to construct prompt configurations for target-domain adaptation. We train a task head with the appropriate output dimension on the target training set while keeping the target pretrained GNN frozen. The baselines follow the same adaptation protocol, transferring and freezing their source-trained prompt parameters or policies while  training the target task head. For each run, all methods use the same training, validation, and test splits on each dataset. We select source checkpoints with the highest source-domain validation accuracy and adapted task-head checkpoints with the highest target-domain validation accuracy. During inference, we retain the prompt-node scheme associated with the selected target checkpoint and use the transferred edge policy to construct connections, with all model parameters fixed. Figure~\ref{fig:cross_dataset} reports the corresponding cross-dataset accuracy.

Across both source settings, TTPR achieves higher accuracy than the baselines on most target datasets, with clear advantages in transfers between citation and co-purchase networks. These results suggest that learned configuration policies can generalize beyond the source domain. Jointly adapting prompt features and connections offers flexibility in both prompt content and propagation, which may help exploit relational information under attribute shifts. RELIEF trails simpler feature-prompting baselines on some targets, suggesting that source-domain dependencies may limit policy transfer and highlighting the potential value of structural adaptation.

\subsection{Time Analysis}

To assess the computational overhead of the proposed two-stage optimization, we compare the average training time per epoch across methods on six datasets using the hardware specified in the experimental settings. Each TTPR epoch includes CRPO, INPPO, and task-head optimization. As shown in Table~\ref{tab:times}, TTPR requires less time per epoch than RELIEF and LEAP on most evaluated datasets, although its advantage varies across datasets. TTPR restricts policy search to prompt-related configurations. CRPO samples a fixed number of complete connection patterns. INPPO incrementally adjusts the prompt count within a bounded range that does not expand with the number of original nodes. This allows a fixed trajectory length independent of input graph size, limiting the number of configuration evaluations. These results show that the two-stage optimization achieves competitive per-epoch training costs relative to the evaluated RL baselines.
}

\subsection{Hyperparameter Sensitivity}

We conduct a sensitivity study on key hyperparameters of TTPR to examine their effects on model stability. Specifically, we examine the maximum number of prompt nodes ($M_{\text{max}}$), the group size $K$ in stage~I, the trajectory length $L$ in stage~II, and the entropy coefficient $\beta$ used in both stages. The analysis is performed under two representative conditions: node classification on the PubMed dataset using the GPPT pre-training strategy, and graph classification on the HIV dataset using the AttrMasking pre-training strategy. 

Figure~\ref{fig:param_analysis} illustrates the effects of several key hyperparameters on performance. For the maximum number of prompt nodes ($M_{\text{max}}$), performance remains relatively stable across a broad range, and near-optimal values can be discovered within each range, indicating the robustness of the framework. For the group size $K$, performance benefits from increasing small groups but shows limited further gains at larger values, suggesting diminishing returns from additional sampled connection patterns. The trajectory length $L$ exhibits a similar pattern, with longer trajectories offering limited additional performance gains beyond a suitable range. Regarding the entropy coefficient $\beta$, intermediate values generally provide favorable performance, consistent with a balance between exploration and task-directed policy learning. These results support choosing suitable hyperparameter ranges for consistent downstream performance.

\begin{table*}[tbp]
\renewcommand{\arraystretch}{1.05}
\centering
\begingroup
\caption{Comparison of All-in-One, PSP, and TTPR in prompt count (\#P), absolute relative density change ($\Delta\rho$), feature magnitude ratio (FMR), and prompt feature similarity (PFS) for 10-shot node classification (CiteSeer and PubMed) and 50-shot graph classification (BACE and BBBP). The three continuous metrics are averaged over test inputs and runs.}
\label{tab:prompt_configuration_analysis}
\fontsize{8pt}{9.6pt}\selectfont
\setlength{\tabcolsep}{3pt}
\begin{tabular*}{\textwidth}{@{\extracolsep{\fill}}l*{12}{c}@{}}
\toprule
Metric & \multicolumn{3}{c}{CiteSeer} & \multicolumn{3}{c}{PubMed} & \multicolumn{3}{c}{BACE} & \multicolumn{3}{c}{BBBP} \\
\cmidrule(lr){2-4}\cmidrule(lr){5-7}\cmidrule(lr){8-10}\cmidrule(lr){11-13}
 & All-in-One & PSP & TTPR & All-in-One & PSP & TTPR & All-in-One & PSP & TTPR & All-in-One & PSP & TTPR \\
\midrule
\#P & 5 & 6 & \textbf{5} & 5 & 3 & \textbf{3} & 5 & 2 & \textbf{4} & 5 & 2 & \textbf{3} \\
$\Delta\rho$ (\%) & 57.80 & 188.40 & \textbf{23.60} & 24.70 & 61.30 & \textbf{9.40} & 62.40 & 132.80 & \textbf{38.70} & 94.60 & 117.50 & \textbf{51.20} \\
FMR & 1.36 & 0.60 & \textbf{1.19} & 0.71 & 0.73 & \textbf{0.86} & 1.24 & 0.85 & \textbf{1.12} & 1.47 & 0.81 & \textbf{0.88} \\
PFS & 0.51 & 0.50 & \textbf{0.39} & 0.41 & 0.83 & \textbf{0.33} & 0.35 & 0.94 & \textbf{0.32} & 0.61 & 0.91 & \textbf{0.46} \\
\bottomrule
\end{tabular*}
\endgroup
\end{table*}

\begin{table*}[tbp]
\makeatletter\def\@captype{figure}\makeatother
\centering
\includegraphics[width=0.96\textwidth]{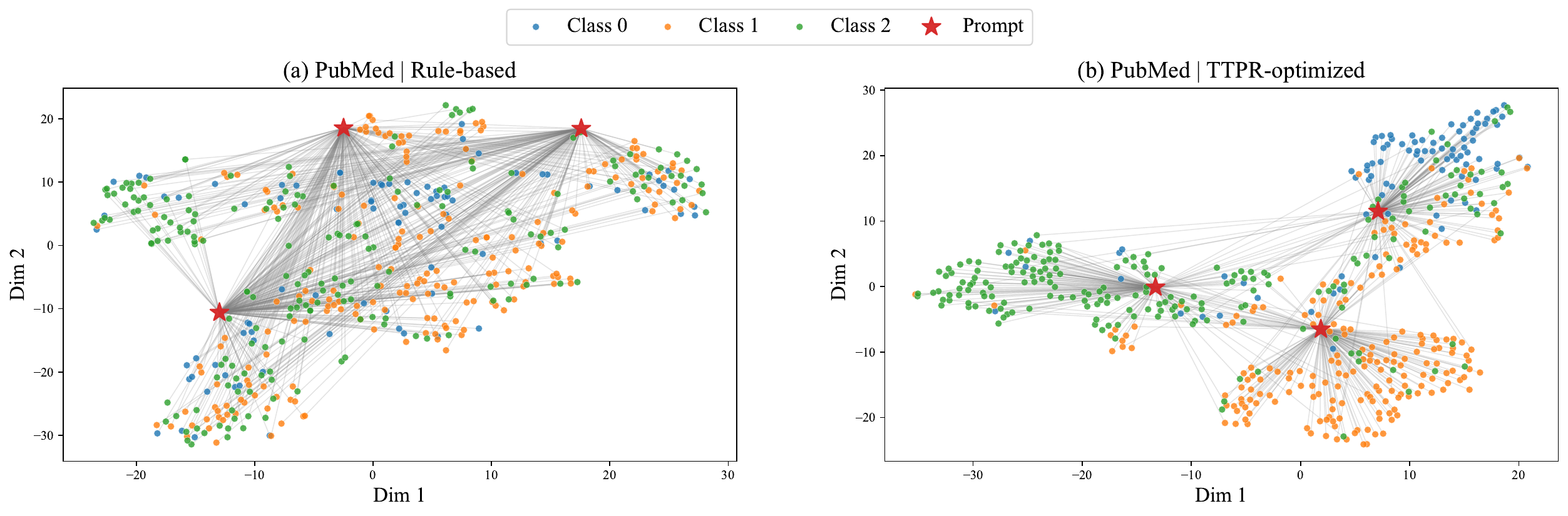}
\caption{Node and prompt embeddings for PubMed after dimensionality reduction under two different configurations.}
\label{fig:case-study}
\end{table*}

\widowpenalty=10000
\subsection{Prompt Configuration Analysis}
\label{sec:prompt_configuration_analysis}

To better understand how TTPR adapts to downstream tasks, we analyze its learned prompt configurations. The prompt count and feature matrix are shared across inputs at the selected checkpoint, while connections are generated for each input by the learned edge policy. For each input with at least one edge, we compute the ratio of augmented to original graph density and express its absolute deviation from one as a percentage. We average these changes across inputs to characterize changes in graph sparsity. Edge density is the fraction of possible unordered node pairs joined by an edge, counting each undirected edge once and excluding self-loops. The augmented graph includes all original nodes, prompt nodes, and their connections. Feature magnitude ratio (FMR) is the ratio of the mean absolute feature values of prompt nodes to those of original nodes, with its deviation from one indicating a mismatch in feature scale. Prompt feature similarity (PFS) is the mean pairwise cosine similarity between distinct prompt feature vectors, evaluated for inputs with at least two prompts. Lower values indicate greater separation between their feature directions. Both feature statistics are computed per input in the continuous space where prompts are injected and then averaged across inputs. We compare All-in-One~\citep{ALL-in-One}, PSP~\citep{PSP2024}, and TTPR on 10-shot node classification with GPPT pre-training on CiteSeer and PubMed, and 50-shot graph classification with AttrMasking pre-training on BACE and BBBP. Node classification statistics cover the input subgraphs of the same 500 test targets, while graph classification statistics cover all test graphs. Both include every original node in each input. Prompt counts are reported as integers, and the remaining metrics are averaged across all five~independent~runs.

Compared with All-in-One and PSP, TTPR induces smaller changes in edge density and keeps FMR closer to one (Table~\ref{tab:prompt_configuration_analysis}), indicating better preservation of graph sparsity and feature scale. Its lower PFS also indicates more distinct feature directions. Together, these properties allow prompts to participate in message passing at scales comparable to those of original nodes while contributing distinguishable feature content. Alongside the performance gains in the main experiments, these observations suggest that TTPR enables effective adaptation through distinct prompt representations while limiting deviations from the original graph structure and feature scale during downstream adaptation.

\subsection{Case Study}

We conduct a case study on PubMed to examine the structural benefits of topology-based prompting configuration learned by our framework. The dataset contains three classes. We compare two configurations: (a) Rule-based, with $M=3$ prompt nodes connected to original nodes whose cosine similarity exceeds $0.7$, and (b) TTPR-optimized, where both the prompt connection patterns and embeddings are learned via our TTPR framework.
For each configuration, we visualize the test-center node representations together with three prompt points in two dimensions, each representing the mean encoded representation of the corresponding prompt node across test subgraphs. As shown in Figure~\ref{fig:case-study}, the Rule-based configuration yields more dispersed node representations in the visualization.
The learned configuration uses a shared scheme with $M=3$ prompt nodes. In the visualization, the prompt points lie near class centers. These anchors act as global structural cues that enhance graph-wide information propagation, promote interactions between related class regions, and improve class separability.

\FloatBarrier
\raggedbottom
\vspace{0pt plus 1fill}
\section{Conclusion}
This paper studies adaptive topology-based prompting for few-shot graph learning, where prompt configurations govern both task information and its structural propagation. Our analysis reveals asymmetric performance effects and coupled configuration selection, making a unified policy difficult to evaluate while preventing the two configurations from being learned independently. TTPR addresses this tension through alternating node-level and edge-level policy learning. CRPO derives edge-level updates from group-wise comparisons of complete connection patterns, whereas INPPO learns prompt-node schemes through incremental decisions with the edge-policy parameters fixed while connections are reconstructed for each updated prompt-node scheme. Across datasets and pre-training strategies, TTPR ranks first in 8/10 node classification settings and achieves the highest average graph classification performance.

\pagebreak
\clearpage
\section*{AI-Generated Content Acknowledgement}
OpenAI ChatGPT and Codex assisted with drafting and revising passages in the Abstract and Sections I--VI, including paragraph organization, language refinement, and terminology consistency. They also assisted with LaTeX formatting and figure and table layout throughout the manuscript, as well as code editing and debugging. The authors retain responsibility for the accuracy and originality of the manuscript and the validity of the reported results.
\bibliographystyle{IEEEtran}
\bibliography{TTPR(arxiv)}

\begin{thebibliography}{10}
\providecommand{\url}[1]{#1}
\csname url@samestyle\endcsname
\providecommand{\newblock}{\relax}
\providecommand{\bibinfo}[2]{#2}
\providecommand{\BIBentrySTDinterwordspacing}{\spaceskip=0pt\relax}
\providecommand{\BIBentryALTinterwordstretchfactor}{4}
\providecommand{\BIBentryALTinterwordspacing}{\spaceskip=\fontdimen2\font plus
\BIBentryALTinterwordstretchfactor\fontdimen3\font minus
  \fontdimen4\font\relax}
\providecommand{\BIBforeignlanguage}[2]{{%
\expandafter\ifx\csname l@#1\endcsname\relax
\typeout{** WARNING: IEEEtran.bst: No hyphenation pattern has been}%
\typeout{** loaded for the language `#1'. Using the pattern for}%
\typeout{** the default language instead.}%
\else
\language=\csname l@#1\endcsname
\fi
#2}}
\providecommand{\BIBdecl}{\relax}
\BIBdecl

\bibitem{pre-train1}
Y.~Xie, Z.~Xu, J.~Zhang, Z.~Wang, and S.~Ji, ``Self-supervised learning of
  graph neural networks: {A} unified review,'' \emph{IEEE TPAMI}, vol.~45,
  no.~2, pp. 2412--2429, 2023.

\bibitem{lowdata1}
Y.~Lu, X.~Jiang, Y.~Fang, and C.~Shi, ``Learning to pre-train graph neural
  networks,'' in \emph{AAAI}, 2021, pp. 4276--4284.

\bibitem{GPL1}
H.~Liu, J.~Feng, L.~Kong, N.~Liang, D.~Tao, Y.~Chen, and M.~Zhang, ``One for
  all: Towards training one graph model for all classification tasks,'' in
  \emph{ICLR}, 2024.

\bibitem{GPL2}
Y.~Li and B.~Hooi, ``Prompt-based zero- and few-shot node classification: {A}
  multimodal approach,'' \emph{CoRR}, vol. abs/2307.11572, 2023.

\bibitem{lowdata3}
Y.~Fang, Y.~Wu, X.~Yu, and S.~Pan, ``Few-shot learning on graphs: From
  meta-learning to llm-empowered pre-training and beyond,'' in \emph{WWW},
  2025, p. 9–12.

\bibitem{Prompt-as-tokens}
Y.~Zhu, Y.~Wang, H.~Shi, Z.~Zhang, D.~Jiao, and S.~Tang, ``Graphcontrol: Adding
  conditional control to universal graph pre-trained models for graph domain
  transfer learning,'' in \emph{WWW}, 2024, pp. 539--550.

\bibitem{GPF}
T.~Fang, Y.~Zhang, Y.~Yang, C.~Wang, and L.~Chen, ``Universal prompt tuning for
  graph neural networks,'' in \emph{NeurIPS}, 2023.

\bibitem{RELIEF}
J.~Zhu, Z.~Ding, J.~Yu, J.~Tan, X.~Li, and W.~Qian, ``{RELIEF:} reinforcement
  learning empowered graph feature prompt tuning,'' in \emph{SIGKDD}, 2025, pp.
  2159--2170.

\bibitem{Prompt-as-graph}
Q.~Huang, H.~Ren, P.~Chen, G.~Krzmanc, D.~Zeng, P.~Liang, and J.~Leskovec,
  ``{PRODIGY:} enabling in-context learning over graphs,'' in \emph{NeurIPS},
  2023.

\bibitem{ALL-in-One}
X.~Sun, H.~Cheng, J.~Li, B.~Liu, and J.~Guan, ``All in one: Multi-task
  prompting for graph neural networks,'' in \emph{SIGKDD}, 2023, pp.
  2120--2131.

\bibitem{PSP2024}
Q.~Ge, Z.~Zhao, Y.~Liu, A.~Cheng, X.~Li, S.~Wang, and D.~Yin, ``{PSP:}
  pre-training and structure prompt tuning for graph neural networks,'' in
  \emph{ECML PKDD}, vol. 14945, 2024, pp. 423--439.

\bibitem{GRPO}
Z.~Shao, P.~Wang, Q.~Zhu, R.~Xu, J.~Song, M.~Zhang, Y.~K. Li, Y.~Wu, and
  D.~Guo, ``Deepseekmath: Pushing the limits of mathematical reasoning in open
  language models,'' \emph{CoRR}, vol. abs/2402.03300, 2024.

\bibitem{PPO}
J.~Schulman, F.~Wolski, P.~Dhariwal, A.~Radford, and O.~Klimov, ``Proximal
  policy optimization algorithms,'' \emph{CoRR}, vol. abs/1707.06347, 2017.

\bibitem{PROMPT}
X.~Sun, J.~Zhang, X.~Wu, H.~Cheng, Y.~Xiong, and J.~Li, ``Graph prompt
  learning: {A} comprehensive survey and beyond,'' \emph{CoRR}, vol.
  abs/2311.16534, 2023.

\bibitem{SUPT}
J.~Lee, W.~Yang, and J.~Kang, ``Subgraph-level universal prompt tuning,''
  \emph{CoRR}, vol. abs/2402.10380, 2024.

\bibitem{GPL4}
Y.~Zhu, J.~Guo, and S.~Tang, ``{SGL-PT:} {A} strong graph learner with graph
  prompt tuning,'' \emph{CoRR}, vol. abs/2302.12449, 2023.

\bibitem{DAGPrompT}
Q.~Chen, L.~Wang, B.~Zheng, and G.~Song, ``Dagprompt: Pushing the limits of
  graph prompting with a distribution-aware graph prompt tuning approach,'' in
  \emph{WWW}, 2025, pp. 4346--4358.

\bibitem{EdgePrompt}
X.~Fu, Y.~He, and J.~Li, ``Edge prompt tuning for graph neural networks,'' in
  \emph{ICLR}, 2025.

\bibitem{DSAS2026}
J.~Li, R.~Wang, Y.~Ma, S.~Liang, G.~Luo, and K.~Qin, ``{DSAS:} {A} universal
  plug-and-play framework for attention optimization in multi-document question
  answering,'' in \emph{NeurIPS}, 2025.

\bibitem{BetaLR2024}
Y.~Ma, K.~Qin, and S.~Liang, ``Beta-{LR}: Interpretable logical reasoning based
  on beta distribution,'' in \emph{NAACL}, 2024.

\bibitem{GraphTOP2025}
X.~Fu, Z.~Lei, Z.~Chen, B.~Zhang, C.~Zhang, and J.~Li, ``{GraphTOP}: Graph
  topology-oriented prompting for graph neural networks,'' in \emph{NeurIPS},
  2025.

\bibitem{UniPrompt2025}
Y.~Huang, J.~Zhao, D.~He, X.~Wang, Y.~Li, Y.~Huang, D.~Jin, and Z.~Feng, ``One
  prompt fits all: Universal graph adaptation for pretrained models,'' in
  \emph{NeurIPS}, 2025.

\bibitem{NeuroPath2025}
J.~Li, R.~Wang, Y.~Huang, Q.~Chen, J.~Zhang, and S.~Liang, ``Neuropath:
  Neurobiology-inspired path tracking and reflection for semantically coherent
  retrieval,'' in \emph{NeurIPS}, 2025.

\bibitem{CLEAR2025}
F.~Wang, Z.~Zhang, J.~Ye, L.~Sun, and J.~Qi, ``{CLEAR}: Cluster-based prompt
  learning on heterogeneous graphs,'' in \emph{PAKDD}, 2025, pp. 264--276.

\bibitem{PLACE2025}
S.~Fang, K.~Zhao, R.~Zhang, Y.~Rong, and J.~X. Yu, ``{PLACE:} prompt learning
  for attributed community search in large graphs,'' in \emph{SIGKDD}, 2026,
  pp. 1039--1050.

\bibitem{GraphToolInstruction2025}
R.~Wang, S.~Liang, Q.~Chen, J.~Zhang, and K.~Qin, ``Graphtool-instruction:
  Revolutionizing graph reasoning in llms through decomposed subtask
  instruction,'' in \emph{SIGKDD}, 2025, pp. 1492--1503.

\bibitem{LEAP}
J.~Xu, Z.~Chen, S.~Yang, J.~Li, H.~Wang, Y.~Li, and E.~C.~H. Ngai, ``Learning
  and editing universal graph prompt tuning via reinforcement learning,'' in
  \emph{SIGKDD}, 2026, pp. 1673--1682.

\bibitem{KEPo2026}
Q.~Chen, C.~Qi, Y.~Huang, M.~Li, R.~Wang, D.~Zhang, K.~Qin, and S.~Liang,
  ``Kepo: Knowledge evolution poison on graph-based retrieval-augmented
  generation,'' in \emph{ACM WWW}, 2026, pp. 2308--2319.

\bibitem{TopoLLM2026}
Y.~Ma, R.~Wang, S.~Liang, G.~Luo, and K.~Qin, ``Topollm: Llm-driven adaptive
  tool learning for real-time emergency network topology planning,''
  \emph{Digital Communications and Networks}, vol.~12, no.~2, pp. 273--282,
  2026.

\bibitem{RL-survey-combination}
N.~Mazyavkina, S.~Sviridov, S.~Ivanov, and E.~Burnaev, ``Reinforcement learning
  for combinatorial optimization: {A} survey,'' \emph{COMPUT OPER RES}, vol.
  134, p. 105400, 2021.

\bibitem{GRL-survey}
V.~Darvariu, S.~Hailes, and M.~Musolesi, ``Graph reinforcement learning for
  combinatorial optimization: {A} survey and unifying perspective,''
  \emph{TMLR}, vol. 2024, 2024.

\bibitem{RL-survey}
G.~P. de~Carvalho, T.~Sawanobori, and T.~Horii, ``Data-driven motion planning:
  {A} survey on deep neural networks, reinforcement learning, and large
  language model approaches,'' \emph{IEEE Access}, vol.~13, pp.
  52\,195--52\,245, 2025.

\bibitem{H-PPO}
Z.~Fan, R.~Su, W.~Zhang, and Y.~Yu, ``Hybrid actor-critic reinforcement
  learning in parameterized action space,'' in \emph{IJCAI}, 2019, pp.
  2279--2285.

\bibitem{RLHF}
N.~Lambert, ``Reinforcement learning from human feedback,'' 2025.

\bibitem{GraphCogent2026}
R.~Wang, S.~Liang, Q.~Chen, Y.~Huang, M.~Li, Y.~Ma, D.~Zhang, K.~Qin, and
  M.~Leung, ``Graphcogent: Mitigating llms' working memory constraints via
  multi-agent collaboration in complex graph understanding,'' in \emph{ACM
  WWW}, 2026, pp. 3811--3822.

\bibitem{RL-for-graph1}
L.~Kong, J.~Feng, H.~Liu, D.~Tao, Y.~Chen, and M.~Zhang, ``{MAG-GNN:}
  reinforcement learning boosted graph neural network,'' in \emph{NeurIPS},
  2023.

\bibitem{RL-for-graph2}
Q.~Sun, J.~Li, H.~Peng, J.~Wu, Y.~Ning, P.~S. Yu, and L.~He, ``{SUGAR:}
  subgraph neural network with reinforcement pooling and self-supervised mutual
  information mechanism,'' in \emph{WWW}, 2021, pp. 2081--2091.

\bibitem{RL-for-graph3}
S.~Hu, Z.~Xiong, M.~Qu, X.~Yuan, M.~C{\^{o}}t{\'{e}}, Z.~Liu, and J.~Tang,
  ``Graph policy network for transferable active learning on graphs,'' in
  \emph{NeurIPS}, 2020.

\bibitem{citation-dataset}
T.~N. Kipf and M.~Welling, ``Semi-supervised classification with graph
  convolutional networks,'' in \emph{ICLR}, 2017.

\bibitem{purchase-dataset}
O.~Shchur, M.~Mumme, A.~Bojchevski, and S.~G{\"{u}}nnemann, ``Pitfalls of graph
  neural network evaluation,'' \emph{CoRR}, vol. abs/1811.05868, 2018.

\bibitem{MoleculeNet}
Z.~Wu, B.~Ramsundar, E.~N. Feinberg, J.~Gomes, C.~Geniesse, A.~S. Pappu,
  K.~Leswing, and V.~S. Pande, ``Moleculenet: {A} benchmark for molecular
  machine learning,'' \emph{CoRR}, vol. abs/1703.00564, 2017.

\bibitem{GPPT}
M.~Sun, K.~Zhou, X.~He, Y.~Wang, and X.~Wang, ``{GPPT:} graph pre-training and
  prompt tuning to generalize graph neural networks,'' in \emph{SIGKDD}, 2022,
  pp. 1717--1727.

\bibitem{GPrompt}
Z.~Liu, X.~Yu, Y.~Fang, and X.~Zhang, ``Graphprompt: Unifying pre-training and
  downstream tasks for graph neural networks,'' in \emph{WWW}, 2023, pp.
  417--428.

\bibitem{masking-contextpred}
W.~Hu, B.~Liu, J.~Gomes, M.~Zitnik, P.~Liang, V.~S. Pande, and J.~Leskovec,
  ``Strategies for pre-training graph neural networks,'' in \emph{ICLR}, 2020.

\bibitem{infomax}
P.~Velickovic, W.~Fedus, W.~L. Hamilton, P.~Li{\`{o}}, Y.~Bengio, and R.~D.
  Hjelm, ``Deep graph infomax,'' in \emph{ICLR}, 2019.

\bibitem{GraphCL}
Y.~You, T.~Chen, Y.~Sui, T.~Chen, Z.~Wang, and Y.~Shen, ``Graph contrastive
  learning with augmentations,'' in \emph{NeurIPS}, 2020.

\end{thebibliography}
\end{CJK*}
\end{document}